\documentclass{svproc}
\usepackage{url}

\usepackage{dingbat}
\usepackage{amsmath}
\usepackage{amssymb}
\usepackage{booktabs}
\usepackage{arydshln}

\usepackage{subcaption}
\usepackage{graphicx}
\usepackage{multirow}
\usepackage{xcolor}
\usepackage{adjustbox} 

\usepackage{mathtools}
\usepackage{bbding}
\usepackage{array}
\usepackage{cite}
\usepackage[numbers,sort&compress]{natbib}

\usepackage[hidelinks,linktoc=all]{hyperref} 
\usepackage[capitalise,nameinlink,noabbrev]{cleveref} 
\usepackage{bookmark} 

\begin{document}
\mainmatter              
\title{Vanishing Depth: Training Generalized Depth Adapters with Sinusoidal Depth Preprocessing for Pretrained RGB Encoders}
\titlerunning{Generalized Depth Adapters with Sinusoidal Depth Preprocessing}  
%
\author{Paul Koch\inst{1} \and Jörg Krüger\inst{1,2}}
\authorrunning{Koch \& Krüger} 
%
%
\institute{Fraunhofer IPK, Pascalstr. 8-9, 10587 Berlin, Germany\\
\email{paul.koch@ipk.fraunhofer.de}\\
\url{https://www.ipk.fraunhofer.de/}
\and
TU-Berlin, Pascalstr. 8-9, 10587 Berlin, Germany\\
\email{joerg.krueger@tu-berlin.de}\\
\url{https://www.tu.berlin/}\\
\texttt{\textbf{Code \& Models}: \url{https://github.com/KochPJ/vanishing-depth}}
}

\maketitle             

\begin{abstract}
Generalized metric depth understanding is critical for precise vision-guided robotics, which current state-of-the-art (SOTA) vision-encoders do not support. To address this, we propose a self-supervised training approach that extends pretrained RGB encoders with a depth adapter to incorporate and align metric depth into a combined latent space without interfering with the pretrained RGB feature extraction. In combination with our sinusoidal depth encoding, the depth adapter enables generalized and robust depth density and distribution invariant feature extraction. Our depth adapters improve a wide set of generalized RGB baselines across a spectrum of relevant RGBD downstream tasks in segmentation, pose estimation, and depth completion -- without the necessity of finetuning. Most importantly, we achieve 56.05 mIoU in the SUN-RGBD segmentation, while outperforming SOTA depth-aware and multi-modal encoders in our experiments. When no depth is present, one can activate our depth adapter with an empty map, use single pixel depth clues, or monocular depth estimation to include the depth aware feature extraction into subsequent downstream tasks. 

\end{abstract}

\section{Introduction}
\label{sec:intro}
Self-supervised learning (SSL) has emerged as a crucial innovation in the development of generalized models, also called foundation models~\cite{FoundationModels}. These models use task-agnostic SSL representation learning to extract generalized features that can be used for downstream tasks without the need for further finetuning~\cite{GPT, DINOv2}. In fact, supervised finetuning might even corrupt generalized pretrained feature extraction~\cite{cripple_foundationmodels, knowledgeinsulation}. The independence from task specific finetuning distinguishes foundation models from traditional transfer learning.  Foundation models yield a faster and more efficient generalized adaptation~\cite{GPT, DINOv2, siglip2, dinov3, knowledgeinsulation}. Moreover, they allow one to append multiple heads for a multi-task system with shared feature extraction~\cite{DINOv2, dinov3}. This further increases computational efficiency, as the encoders are often the largest modules in AI-based computer vision and enable fast modular updates. We observe in the literature a general push towards foundation models that are capable of processing multiple input modalities~\cite{multimae, OmniVore, OmniSegmentor, OmniVec, OmniVec2}. Within computer vision, MultiMAE~\cite{multimae}, Omnivore~\cite{OmniVore}, and OmniVec~\cite{OmniVec, OmniVec2} use multi-modal pretraining to create generalized multi-modal capable vision encoders with promising properties. LLaVa~\cite{llava}, Clip~\cite{clip}, and EVA~\cite{EVA, eva02} employ text-image encoders that yield SOTA results. However, most related research on generalized vision encoders focuses primarily on RGB~\cite{MAE, iBot, DINOv2, dinov3, eva02}. Here, we observe a lack of vision encoders with metric depth understanding in the literature, which seems especially crucial for downstream tasks that require absolute 3D position information, particularly in the field of robotics~\cite{knowledgeinsulation}, where VLA~\cite{openvla, pi0_paper} or robot policy learning~\cite{diffusionPolicy} image encoders generally benefit from pretrained encoders~\cite{Unsurprising}, but require finetuning~\cite{diffusionPolicy, pi0_paper, openvla}. 

In this work, we propose a depth adapter that extends pretrained RGB encoders, like DINOv2~\cite{DINOv2}, to extract and incorporate depth features into an aligned latent space. Through this approach, we demonstrate a generalized metric depth-aware RGBD encoder capable of addressing various RGBD downstream tasks without the need for additional finetuning. In addition, we introduce sinusoidal depth preprocessing (SDP), which, in our experiments, exhibits stable scalability when applied to novel depth distributions, densities, and perturbations, unlike commonly used depth normalization approaches. Our methods, in combination with DinoV2, result in a generalized RGBD encoder that produces better results across a wide range of downstream tasks in segmentation, depth completion, and pose estimation. We avoid VLA and policy learning tasks in our ablation studies for now, as their benchmarking and fair comparison are not yet working out of the box, and real world experiments on robots are not within the budget of this work. 

\section{Related work}
\label{sec:related_work}

\textbf{Pretrained depth encoders} have demonstrated their value in solving RGBD downstream tasks (e.g. ~\cite{DFormer-semi-ssl-cls, clipdepth, rasim}. However, these encoders are often tailored for specific tasks or datasets and lack the generalizability offered by, e.g. DINOv2~\cite{DINOv2} and DINOv3~\cite{dinov3}. However, SOTA methods for RGBD segmentation often leave the depth encoder randomly initialized, relying solely on supervised training~\cite{tokenfusion, cross-modal-fusion, PanopticNDT, GeminiFusion}. Other related work employs SSL methods such as inpainting~\cite{ssl-depth-inpainting} and combinations of monocular depth estimation with recolorizing~\cite{ssl-detection}. Others explore new depth SSL options such as modality-fusion~\cite{SSL-multi-modal}, segmentation label generation~\cite{SSL-depth2RGBseg}, and 3D-predictions~\cite{ssl-depth-2-3d}. However, these methods are often task specific, do not scale towards broad applications, and do not show depth density, distribution, or noise invariance. DFormerV1~\cite{DFormer-semi-ssl-cls} and DFormerV2~\cite{DFormerv2} use classification based pretraining of their RGBD encoders, yielding SOTA results in segmentation. 

\textbf{Multi-Modal (RGBD) encoders} aim to create feature extraction that can handle any given input modality and support a wide range of downstream tasks. Encoders such as MultiMAE~\cite{multimae}, Omnivore~\cite{OmniVore}, Omnivec~\cite{OmniVec}, OmniVec2~\cite{OmniVec2}, and OmniSegmentor~\cite{OmniSegmentor} use token-based fusion and align the modality wise feature extraction within a common latent space. While Omnivore and OmniSegmentor use supervised multi-task classification and segmentation pretraining, respectively, OmniVec uses contrastive MiM~\cite{mim} to pretrain their encoders, while OmniVec2 uses a discriminative SSL approach (see, e.g., Dino~\cite{DINOV1}). CLIP2Point~\cite{clipdepth} uses contrastive learning to align depth encoding with CLIP~\cite{clip}. However, these training methods do not explicitly guide the understanding of metric depth. Instead, they align the general context of the input modalities without imposing pressure on pixel-wise depth precision. This proves fine for classification and segmentation purposes~\cite{OmniVore, OmniVec, OmniVec2, clipdepth, OmniSegmentor, DFormerv2}, but we demonstrate this inaccuracy in our depth completion and pose estimation experiments, where pixel-wise metric depth precision is more important (see Fig.~\ref{fig:radar_plots}\&\ref{fig:depth_noise_curves}). 

\textbf{Depth Encoding and Completion:} In contrast to these aforementioned methods, SSL MAE-like~\cite{MAE} RGB-guided depth completion~\cite{DepthCompletion5, DepthCompletion1, DepthCompletion2, DepthCompletion3, DepthCompletion4, DepthCompleteion_SEMATTNET, rasim} uses the semantic RGB information to guide the sparse to dense prediction of missing depth values. This trains both semantic understanding and precise metric depth feature extraction but ignores explicit classification-specific feature extraction. RaSim~\cite{rasim} shows a good understanding of metric depth and generalizes to robotic applications but fails to account for depth invariance at inference time. 

\textbf{Depth Understanding:} A different branch of related work avoids encoding depth and instead trains their RGB encoders to have inherent depth understanding. In generalized monocular depth estimation, such as Depth Anything V2~\cite{depthanythingv2} and UniDepth~\cite{unidepth}, one can observe a comprehensive zero shot depth map estimation. Other works such as Dune~\cite{dune}, Fit3D~\cite{fit3d}, MASt3R~\cite{MASt3R}, or CroCo V2~\cite{crocov2} focus on including depth in their pretraining objectives to enrich the feature space of the RGB encoder with metric depth understanding. This results in improved depth aware encoders with relative gains on related downstream tasks. 

\textbf{General purpose models} are a set of pretrained model architectures that can support a wide range of downstream problems without the need for finetuning. In computer vision, we use SSL-pretrained encoders such as Dino V1-3~\cite{DINOV1, DINOv2, dinov3}, MAE~\cite{MAE, multimae}, EVA~\cite{EVA, eva02}, SigLip2~\cite{siglip2}, or Clip~\cite{clip, clipdepth}, which yield generalized vision embeddings extracted via general encoder architectures such as ViTs~\cite{ViTs}, CNNS~\cite{resnet}, or Swin~\cite{swin}. These encoder architectures extract hierarchical feature maps and descriptive tokens that can be used by subsequent decoder heads to address any downstream vision task. Unlike pure RGB image encoders (e.g., DinoV2), some are text aligned (e.g., EVA~\cite{EVA}, SigLip2~\cite{siglip2}, Clip~\cite{clip}, or DINOV3~\cite{DINOv2}), while others are Multi-Modal vision encoders (e.g., MultiMae~\cite{multimae}, OmniVore~\cite{OmniVore}, or OmniSegmentor~\cite{OmniSegmentor}). Language model driven VLMs~\cite{introvlm} and VLAs~\cite{openvla, pi0_paper} leverage these pretrained generalized encoder modules to gain generalized image understanding. However, it has been found that vision encoders have a gap, especially in transferability to robotic applications~\cite{knowledgeinsulation}. We further find that these encoders also exhibit a gap in general precise metric depth understanding. We aim to reduce this gap with our adapters, which can be activated on existing pretrained encoders.

\section{Depth Adapter (DA)}
\label{sec:methods}

\begin{figure}[h!]
    \centering
    \includegraphics[width=1.0\linewidth]{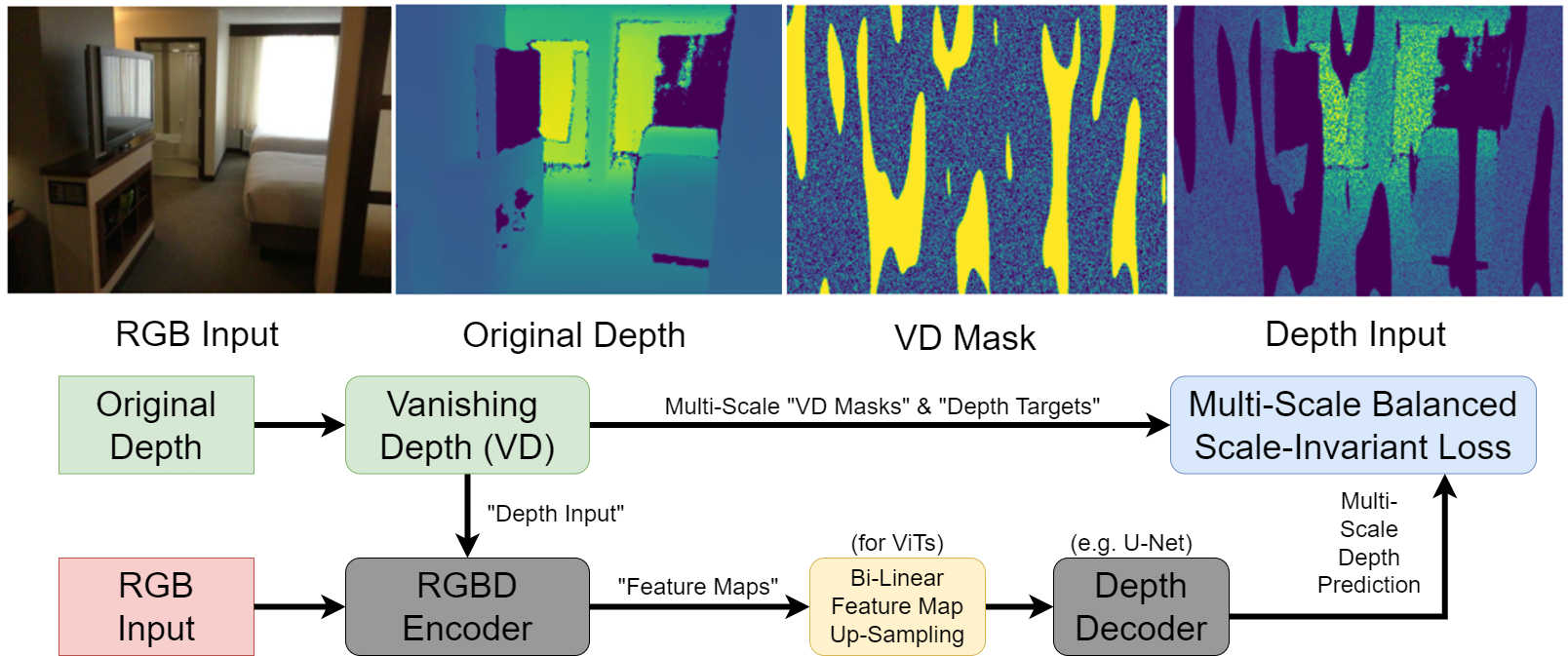}
    \caption{\textbf{Vanishing Depth Training Pipeline:} We use a mixture of perlin and random resized noise to remove depth pixels from the original depth image. The remaining depth map is than passed through an RGBD encoder and FPN decoder network~\cite{FPN}. Using multi-scale noise masks, we use scale-invariant loss for reconstructing the input depth and predicting missing depth inputs at multiple downstream stages.}
    \label{fig:DAPipeline}
\end{figure}
\begin{figure}[h!]
    \centering
    \includegraphics[width=0.5\linewidth]{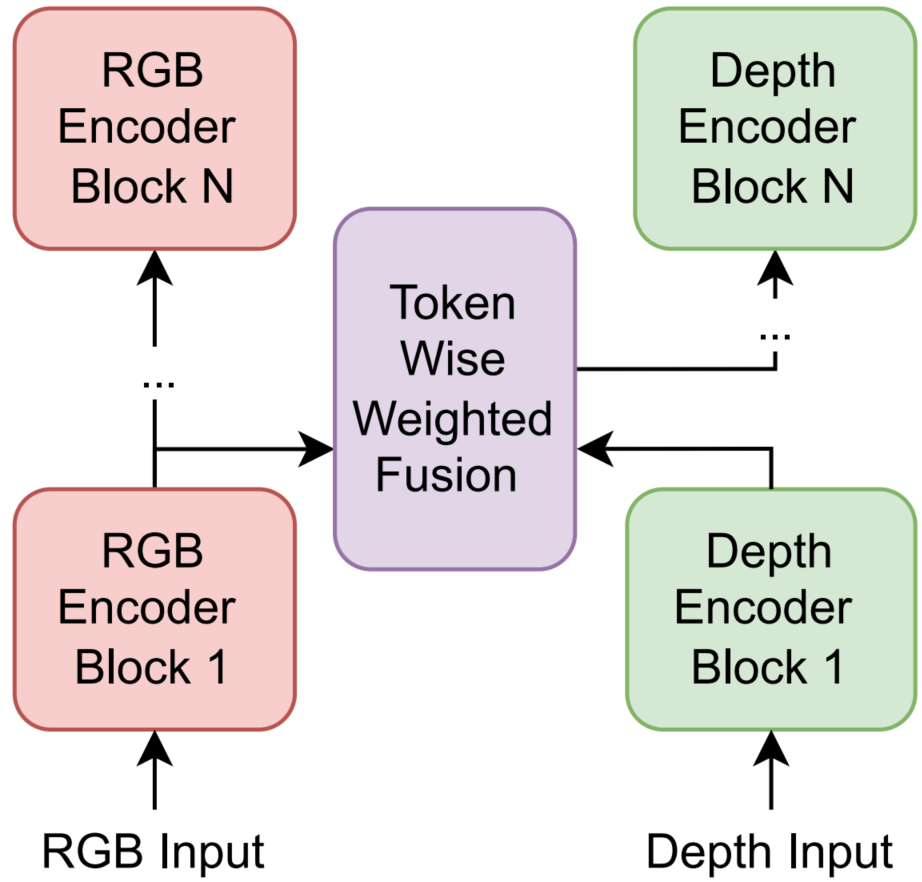}
    \caption{\textbf{RGBD Encoder Architecture:} One can use any commonly used encoder (ViT, CNN, Swin, etc) independent between RGB and Depth or size.}
    \label{fig:rgbd_encoder}
\end{figure}

Our vanishing depth pretraining pipeline (see Fig.~\ref{fig:DAPipeline}) for depth adapters is inspired by previous work in in-painting~\cite{inpainting-SSL, ssl-depth-inpainting}, monocular depth estimation~\cite{scale-invariance, Monocular-depth-ssl}, color-guided depth completion~\cite{DepthCompletion3, rasim}, and ViT Adapters~\cite{vitadapter} for dense metric depth prediction. We combine these into our depth adapter to create a generalized RGBD encoder that can be used for a wide range of unseen downstream tasks, without training an RGBD encoder from scratch or finetuning. The depth adapter leverages existing pretrained RGB encoders to benefit from their generalized feature extraction. The extracted color features should be fused into depth embeddings at various stages in the depth encoder latent space. This leaves the pretrained RGB encoder unspoiled and allows the depth encoder to incorporate hierarchical RGB embeddings. 

\subsection{Implementation}
Following generalized RGB vision encoders~\cite{eva02, DINOv2}, we use ViTs~\cite{ViTs} for both our vision encoders. We use ViT-B as most methods support this model size, and larger models should not meet the high frequency and low latency requirements in robotics. For gradual intermediate RGBD fusion (see Fig.~\ref{fig:rgbd_encoder}), we use a simple S\&E~\cite{squeeze-and-excitation} to combine RGB and depth signals at the layers [3, 6, 9, 12] (similar to~\cite{DPT}). Afterwards, the fused RGBD signals are forwarded to the subsequent depth layer, respectively. Note that any other combination of RGB or depth encoder architectures can be used independent and across sizes (small, base, large, etc.), fusion stages (layers or blocks), or types (CNNs~\cite{resnet}, ViT~\cite{ViTs}, or Swin~\cite{swin}). With S\&E fusion, one can change feature dimensions while interpolations enable shape adaptations~\cite{vitadapter, DPT}. Thus, the count of model parameters varies depending on the architecture design. With our $2\times$Vi-B$=2\times86$mio, we have approximately twice the parameters of a normal ViT-B, which also doubles inference time. However, this allows us to initialize both the RGB encoder and the depth encoder with the same pretrained RGB encoder weights (all but the first depth encoder layer, depending on the input dimension). This is training efficient since the RGB embedding is fused into the depth encoder for guidance, and thus, the depth encoder gets a head start. Again, one could also combine an RGB ViT-B with a depth ViT-S to make the adapter more lightweight, or use model distillation~\cite{model_distilling, DINOv2} for non pretrained depth encoder architectures. During training, we freeze the pretrained RGB encoders, which allows us to leverage the pretrained RGB encoder without corrupting it~\cite{cripple_foundationmodels}. For pixel-wise decoding, we once again use the RGBD fused attention maps at layers [3, 6, 9, 12] and bilinear resize them for a simple and lightweight U-Net decoder~\cite{Unet} with four FPN~\cite{FPN} heads for multi-scale decoding (again, any other dense decoder is viable, e.g., DPT~\cite{DPT}). For metric depth decoding, we use a lightweight implementation inspired by Adabins~\cite{Adabins}. Given the max depth $\max(d)$, we decode a vector of length $n$ per depth index $j$ from which we compute the final depth $d_j$ by $d_j = f(\sum_i^n \frac{\max(d)}{10i})$, where $\frac{\max(d)}{10n} > 1mm$ and $f$ are leaky ReLU functions to avoid and recover from negative outcomes. This stabilizes the decoding as decoded values are mostly within $\pm1$.

\subsection{Sinusoidal Depth Preprocessing (SDP)}
\label{sec:encodedepth}
\begin{figure*}[h]
\centering
\begin{subfigure}[t]{0.16\textwidth}
     \centering
     \includegraphics[width=\textwidth]{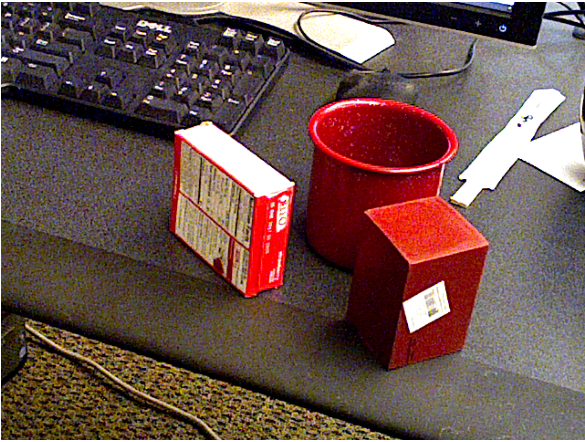}
     \caption{RGB}
\end{subfigure}
\begin{subfigure}[t]{0.16\textwidth}
     \centering
     \includegraphics[width=\textwidth]{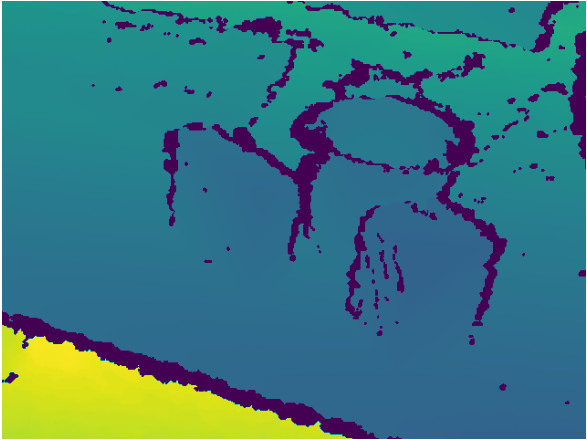}
     \caption{Depth}
\end{subfigure}
\begin{subfigure}[t]{0.16\textwidth}
     \centering
     \includegraphics[width=\textwidth]{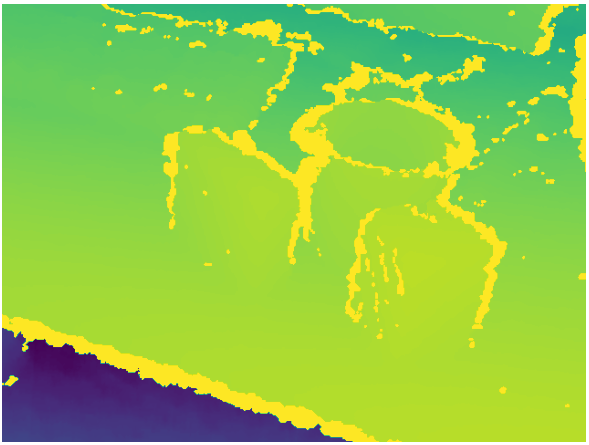}
     \caption{1 ch.\ \\ 
     $(\approx15.0m)$}
\end{subfigure}
\begin{subfigure}[t]{0.1565\textwidth}
     \centering
     \includegraphics[width=\textwidth]{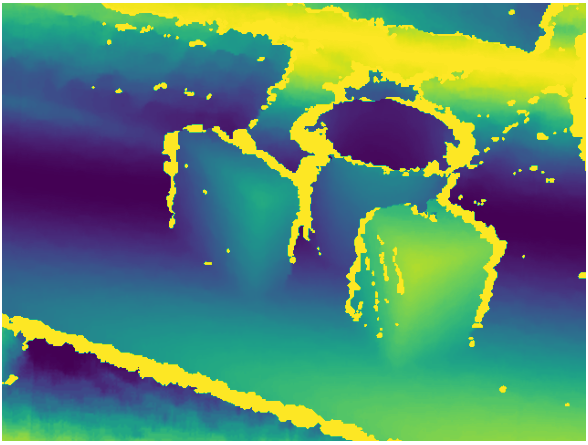}
     \caption{10 ch.\\ 
     $(\approx1.97m)$}
\end{subfigure}
\begin{subfigure}[t]{0.16\textwidth}
     \centering
     \includegraphics[width=\textwidth]{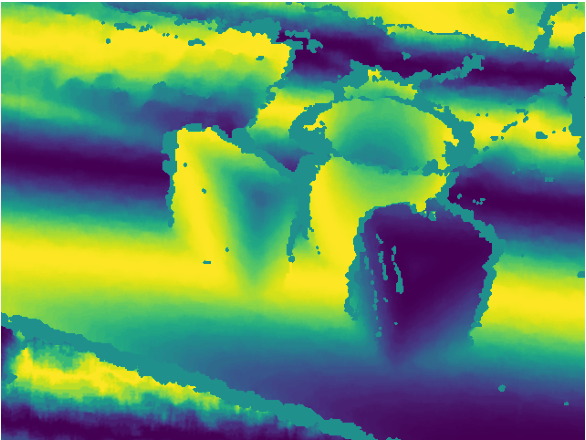}
     \caption{16 ch.\\ $(\approx0.43m)$}
\end{subfigure}
\begin{subfigure}[t]{0.16\textwidth}
     \centering
     \includegraphics[width=\textwidth]{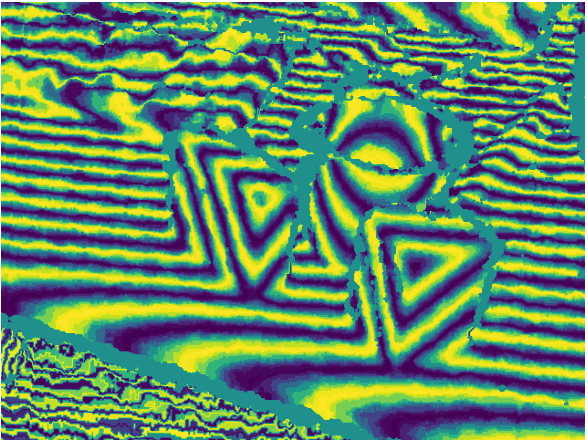}
     \caption{32 ch.\\ $(\approx0.0075m)$}
\end{subfigure}
\caption{\textbf{Visualization of SDP:} Example of SDP with 32 channels (ch.) and max depth $15m$. The first frequency (c) captures large changes in depth image, while the final frequency (f) encodes minor depth changes.}
\label{fig:SDP}  
\end{figure*}

A RGBD encoder should be able to handle various input distributions, densities, and related downstream tasks under depth perturbations, which we found the current depth preprocessing methods struggle to yield. Therefore, we introduce sinusoidal depth preprocessing (SDP) for stable feature extraction from continuous real numbers:
\begin{equation}
  \begin{aligned}
    l = 2\pi (d_j/\max(d)) \\
    SDP(d_j, 2i) = \sin(l/T^{2i/c}) \\
    SDP(d_j, 2i+1) = \cos(l/T^{2i/c}) \\
\end{aligned}
\end{equation}
where $i$ is the SDP channel and $d_j$ is the depth pixel of the depth map $d\in D$, and $D$ is the expected depth distribution. $c$ the number of depth channels in SDP and $T<1$ is the temperature that defines the frequency distribution of SDP.  
\begin{equation}
\max(d) = 
\begin{cases}
    \max(D),& \text{if using global max depth (GSDP) }\\
    \max(d),              & \text{otherwise local max depth (LSDP)}
\end{cases}
\end{equation}
normalizes the input depth. We use global (GSDP) and local (LSDP) to distinguish between the two SDP-modes. If using LSDP, the model needs to encode $\max(d)$ to be able to decode metric depth. Otherwise, with GSDP the model can learn the constant global max scalar during training. For our ViTs~\cite{ViTs} with LSDP, we encode $\max(d)$ into a vector $V$ and add it to the depth encoder's CLS-token. We vectorize $\max(d)$ by encoding its integer and decimal values $v$ using $V_i = (v_i+1) / 10$. We pad the vector with zeros to fit the size of the CLS-token and hold the separation index between integer and decimal values constant at index 64. This enables the model to read any $\max(d)$ with up to 64 integer and decimal places. SDP yields stable input and feature extraction within the range of -1 and 1, regardless of the input distribution or absolute depth (see Fig.~\ref{fig:SDP}). However, SDP increases the computational cost linearly by a factor of $c$ in the first convolution or fully-connected encoding layer, and limits the encoding precision to the distribution ($T$) and the number of encoding frequencies ($c$). SDP differs from classical sinusoidal positional encoding (PE)~\cite{attention-is-all-pe} used in transformers in that it is employed as a preprocessing method for continuous numbers, and allows dynamic input normalization that can be decoded with the correct key into metric depth. Moreover, we use $T<1$ to define the smallest frequency, which is more intuitive for depth. SDP is not comparable to other PE methods (e.g., RoPe~\cite{RoPE} or learnable PE~\cite{attention-is-all-pe}), as it serves the preprocessing of continuous numbers with a completely different purpose than the token localization in attention models of PE. 

\subsection{Vanishing Depth Training Pipeline Toolbox}
The following list of methods used in our training pipeline (see Fig.~\ref{fig:DAPipeline}) is designed to vanish any connection between RGB and metric depth while mixing the input depth density and distribution. This allows us to train a depth density and distribution invariant stable feature extraction with a generalized semantic depth understanding based on metric and random non-metric depth clues, while avoiding any unwanted RGB-based monocular depth estimation or classification from occurring. 

\textbf{Distribution invariance:} A key factor in our pretraining pipeline is to randomly manipulate the depth images by rescaling and offsetting them. This prevents monocular depth estimation from the RGB encoder, as each target depth training sample can have numerous variations. Thus, the model has to combine semantic RGB with depth embeddings. We employ random offsets, random depth scalars, and rescaling of the depth image within specific minimum and maximum values drawn from jittered bins. The bins ensure that larger numbers are not overrepresented. We randomize our input depth with $\frac{1}{4}$ random depth jitter, $\frac{1}{4}$ rescales between two randomly selected and sorted bins, and $\frac{1}{4}$ random offsets. The remaining $\frac{1}{4}$ remains at the original metric depth. The jitter reduces or increases the original depth samples by $\pm 20\%$. The offsets are drawn uniformly per depth sample $d$ from its $\pm STD(d)$. The bins are randomly drawn from $0-1m$, $0.5-2m$, $0.5-5m$, $1-15m$, $5-max_d$ and jittered as well. The depth samples exceeding the maximum training depth $max_d$ are rescaled to $max_d \times (0.9+ r0.1)$, where $r$ is some random scalar. This corresponds to a training distribution of $\approx21.59\pm5.82$ for $max_d = 500$m and $\approx2.84\pm.79$ for $max_d = 15$m. 

\textbf{Smooth and ignore noisy depth:} We also apply random noise during training to sensitize our depth encoders to measurement errors. We add small gaussian noise of $5$~mm to smooth out fluctuations and occasionally also add large random noise between zero and the maximum depth to sensitize the encoder to ignore significant measurement errors. 

\textbf{Density invariance:} To train inpainting models, binary masks or similar markings are required to indicate the areas to be removed. Some approach this with masked grids~\cite{MAE, ssl-depth-inpainting, inpaint3, inpaint4} or generate random noise with a threshold from a given distribution to obtain the binary mask. This can be applied to different scales of the input size, which are then up-sampled to control the size of the removed areas. Another approach is to use perlin noise, which provides random noise with rounded organic shapes that vary in size. We use both approaches to: A) mimic sparse lidar depth sensors with random noise; We remove between 1–99\% of the depth input, enabling the depth encoder to handle any depth density.  

\textbf{Multi-scale balanced scale-invariant loss:} The scale-invariant loss introduced by \cite{scale-invariance} is commonly used in depth completion. Traditionally, this loss is applied to predicted depth values where no input is present. However, our objective is to be equally effective at reconstructing given input pixels and predicting missing input pixels. Since we further randomize the depth input density, we balance our loss between each sample's masked scale-invariant reconstruction and prediction loss. This stabilizes training, as both the easy depth reconstruction and the hard depth prediction become equally important, regardless of the depth density distribution in the training batch. Moreover, we use multi-scale depth decoding to ensure hierarchical depth encoding and RGB feature alignment throughout the depth encoder, which might otherwise be compromised due to skip connections for the decoder. The loss is defined as:
\begin{equation}
    \mathcal{L}_{SI} = \alpha \sqrt{\frac{1}{T}\underset{i}\Sigma g^{2}_{i} - \frac{\lambda}{T^2} (\underset{i}\Sigma g_i)^2 }
\end{equation}
where $g_i = \log \hat{d_i} - log d_i$, $\hat{d_i}$ is the depth prediction at pixel index $i$ and $d_i$ is its ground truth (pixels without depth ground truth are ignored). The constants $\lambda = 0.85$ and $\alpha = 10$ are used to balance the two loss terms~\cite{Adabins}. Given our vanishing depth mask $\mathcal{M}$ (the removed input depth pixels), we use the SI-loss twice; first (1) on pixel wise depth reconstruction ($i \notin \mathcal{M}$), and after (2) on pixel wise depth estimation ($i \in \mathcal{M}$):
\begin{equation}
    \mathcal{L}_{pixel} = \mathcal{L}_{SI}(i \notin \mathcal{M}) + \mathcal{L}_{SI}(i \in \mathcal{M})
\end{equation}
The final "multi-scale" (MS) part of our loss is simply computing $\mathcal{L}_{pixel}$ at multiple output resolutions:
\begin{equation}
    \mathcal{L}_{MS} = \underset{\mathcal{F}}\Sigma  \mathcal{L}_{pixel}(\mathcal{R}(\mathcal{M}, \hat{d}, d, \mathcal{F}))
\end{equation}
where $\mathcal{R}$ is a resize function given some resize factor $\mathcal{F}$ to fit the shape of the ground truth to the decoder shape at each given stage. 

\begin{table*}[h]
\small
  \centering
  \caption{\textbf{Training and Test Datasets:} We report the image count, dataset objective, and mean$\pm$std and proportion of missing depth.}
  \label{tab:datasets}
  \begin{tabular}{@{}l|cclccc@{}}
    \toprule
    &Dataset Name & RGBD-Images &\multicolumn{2}{c}{Objective} &\multicolumn{2}{c}{Depth [m] @ density}\\
    \midrule
    \parbox[c]{3mm}{\multirow{9}{*}{\rotatebox[origin=c]{90}{Training Data}}}&Cityscapes~\cite{CityScapes}& ~104k & \multicolumn{2}{c}{City Segmentation}& 27.7$\pm$26.0& 78\%\\
    &NYU-v2~\cite{nyuDepthV2}&482k &\multicolumn{2}{c}{Indoor Segmentation}&4.1$\pm$1.5& 86\%\\
    &YCBV~\cite{PoseCNN} &113k & \multicolumn{2}{c}{Object 6D-Pose}&1.2$\pm$0.5&82\%\\
    &LM~\cite{LineMod} & 50k  & \multicolumn{2}{c}{Object 6D-Pose}&1.1$\pm$0.4&90\%\\
    &T-Less~\cite{T-Less} &50k & \multicolumn{2}{c}{Object 6D-Pose}&1.0$\pm$0.4& 92\%\\
    &GraspNet~\cite{GraspNetDs} &51  &\multicolumn{2}{c}{Robot Grasping-Pose}&0.5$\pm$0.2&90\%\\
    &ScanNet~\cite{ScanNet} & 2 477k & \multicolumn{2}{c}{Indoor Segmentation} & 1.8$\pm$0.5& 100\%\\
    &MatterPort3D~\cite{Matterport} & 194k & \multicolumn{2}{c}{Indoor Segmentation} & 2.0$\pm$0.7&100\%\\
    &BlendedMVS~\cite{BlendedMVS} & 48k & \multicolumn{2}{c}{Multi-View Stero}&83.4$\pm$24.7&100\%\\
    \midrule
    \parbox[c]{3mm}{\multirow{6}{*}{\rotatebox[origin=c]{90}{Test Only}}}&Kitti~\cite{kitti} & 7k &Depth Estimation& & 15.7$\pm$11.5 & 16\%\\
    &SUN-RGBD~\cite{SUNRGBD} & 5k & \multicolumn{2}{c}{Indoor Segmentation}&2.3$\pm$0.7&87\%\\
    &Void~\cite{VoidDataset}&  59k & \multicolumn{2}{c}{Depth Completion} & 1.65$\pm$0.99 & 96\% \\ 
    &HomeBrew~\cite{HomeBrew} & 50k & \multicolumn{2}{c}{Synt2Real 6D-Pose} & 1.25$\pm$0.94 & 84\% \\
    &MVIP~\cite{MVIP}& 37k& \multicolumn{2}{c}{Industrial Objects} & 0.91$\pm$0.88 & 85\%\\
    &LM-O~\cite{LM-O}&50k  & \multicolumn{2}{c}{Object 6D-Pose}&1.1$\pm$0.4&90\%\\
    \bottomrule
  \end{tabular}
\end{table*}

\subsection{Ablation Study}
 We train our depth adapters with the datasets listed in Table~\ref{tab:datasets} for 100 epochs using our multi-scale balanced scale-invariant loss, which we balance between depth reconstruction and prediction at every output stage. Each epoch consists of 20k samples drawn evenly from the datasets. We use adam optimization with a batch size of $128$ at a resolution of $224$ and a learning rate of 1e-5. We employ augmentations of color jitter, random grayscale, flips, and rotation in combination with our vanishing depth training toolbox.
 
\begin{figure}[h]
    \centering
    \includegraphics[width=1.0\linewidth]{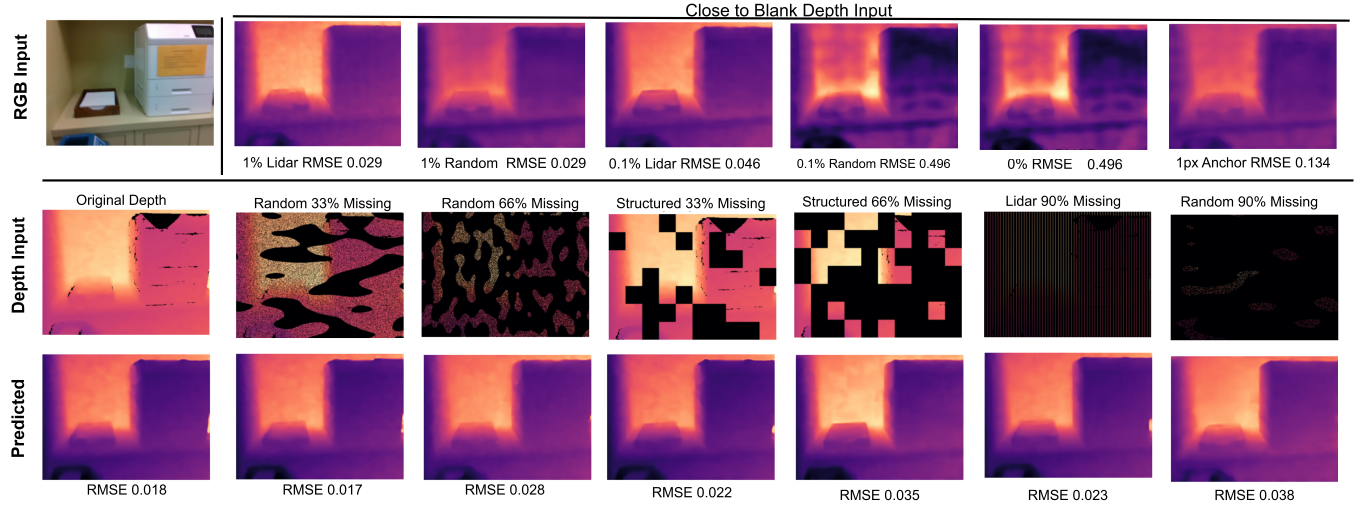}
    \caption{\textbf{Visualization of Depth Prediction}: Our depth adapters with SDP yield stable output with different depth input densities and noise functions mimicking random, structured, sparse and lidar depth inputs. The DA can also function with a single pixel depth anchor or even with a zero map input, generating its own depth input or working without a real depth map at all.}
    \label{fig:qualtydepth}
\end{figure}

\begin{table}[h]
\small
  \centering
  \caption{\textbf{Ablation Study:} We compare our depth adapters and SDP vs. related work on depth encoding; normalization (Norm) with some distribution ($2.4\pm.6$ dataset mean, $5\pm5$ covers $15$~m reasonably, and norm disparity maps~\cite{OmniVore} (Norm Disp.). We report the mean AUC RMSE (1:5:96\% depth). \tiny{\eye}\normalsize seen in DA-training.} 
  \label{tab:DA_results}
  \begin{tabular}{l l | lr |cc | c }
    \toprule
    &\multicolumn{2}{c}{}&Task~$\rightarrow$&\multicolumn{3}{c}{Depth Completion (lower is better)}\\
    &\multicolumn{2}{c}{}&Metric$\rightarrow$&\multicolumn{3}{c}{mean AUC RMSE [mm] }\\
    &\multicolumn{2}{c}{}&Dataset$\rightarrow$&Citys.\tiny{\eye}&NYU.\tiny{\eye}&Void\\
    &\multicolumn{2}{c}{}&Depth$\rightarrow$ &27.7$\pm$26.0& 4.1$\pm$1.5& 1.65$\pm$0.99\\
    \midrule   
    \parbox[c]{3mm}{\multirow{11}{*}{\rotatebox[origin=c]{90}{Ablation Study [15m]}}}&
    \parbox[c]{3mm}{\multirow{11}{*}{\rotatebox[origin=c]{90}{(Ours)}}}&\multicolumn{2}{l|}{Norm Disp.}& 3765.0 & 139.1 & 63.0\\
    &&\multicolumn{2}{l|}{Norm 2.4$\pm$.8}& 3761.7 &133.4 & 63.2\\
    &&\multicolumn{2}{l|}{Norm 5$\pm$5}& 3652.6 & 129.8 & 61.7\\\   
    &&\multicolumn{2}{l|}{LSDP-64c}& 3024.7 & 97.4 & 43.6\\
    &&\multicolumn{2}{l|}{LSDP-32c}& 3060.1 & 101.0 & 45.4\\
    &&\multicolumn{2}{l|}{GSDP-64c}&\textbf{2995.6} & \textbf{94.4} & \textbf{40.2}\\
    &&\multicolumn{2}{l|}{GSDP-32c}& 3041.8 & 95.3 & 40.5\\       
    &&\multicolumn{2}{l|}{~~~~ w. DinoV3} & 3366.8 & 106.1 & 45.5\\
    &&\multicolumn{2}{l|}{~~~~ w. EVA02} & 3216.4 & 102.6 & 42.3\\
    &&\multicolumn{2}{l|}{~~~~ no Perlin Noise} & 2877.2 & 86.7 & 35.8\\ 
    &&\multicolumn{2}{l|}{~~~~ no Pretrained} & 3548.4 & 119.1 & 52.0\\
    \midrule      
    \parbox[c]{3mm}{\multirow{4}{*}{\rotatebox[origin=c]{90}{[$500m$]}}}&
    \parbox[c]{3mm}{\multirow{4}{*}{\rotatebox[origin=c]{90}{[(Ours)]}}}
    &\multicolumn{2}{l|}{Norm 21$\pm$6}&3840.8 & 156.8 & 82.3\\     
    &&\multicolumn{2}{l|}{LSDP-16c}&3145.6 & 120.3 & 60.2  \\
    &&\multicolumn{2}{l|}{LSDP-32c}&2974.0 & 113.8 & 59.4\\
    &&\multicolumn{2}{l|}{LSDP-64c}& \textbf{2940.9} & 113.4  & 61.9\\
    \midrule   
    \parbox[c]{3mm}{\multirow{2}{*}{\rotatebox[origin=c]{90}{SOTA}}} & 
    \parbox[c]{3mm}{\multirow{2}{*}{\rotatebox[origin=c]{90}{DC}}}
    &\multicolumn{2}{l|}{DM3C}& 4733.4 & 200.4 & 126.5\\
    &&\multicolumn{2}{l|}{OmniDC}& 4048.9 & 89.0 & 44.2\\
    \bottomrule
  \end{tabular}
\end{table} 
For comparison, we train multiple baselines, each with different norm-based depth preprocessing, and report the mean AUC of the commonly used RMSE metric on multiple datasets (see Table~\ref{tab:DA_results}). We compare these with different SDP implementations, each with individual $\max(d)$, channels, and pretrained RGB weights. We investigate $c=64$ and compare it with the $32$ and $16$ channel versions, which are computationally lighter but less precise. In addition, we perform experiments with different SOTA pretrained encoders and for $\max(d) = Max(D)$ of $15$m. We selected $15$m because most indoor depth cameras are bound to around that number, and $500$m for extreme outdoor tasks. The results show that our SDP-based encoders yield the best performance with respect to a set of sparse to dense depth completion tasks. Moreover, while $\max(d) = Max(D)$ achieves the best results at large distances, we find that fixing the SDP to $15m$ and rescaling large distances is nearly as good. Similarly, we see that the $32$ channels are sufficient for SDP-based encoding in the $15m$ range. For further qualitative assessment of our methods, we visualize the attention maps of our best model in Fig.~\ref{fig:vit_embs} and depth reconstruction in Fig.~\ref{fig:qualtydepth}. 
\def\spacerDE{0.185}
\def\spacerDET{0.014}
\begin{figure*}[h]
\centering
\begin{subfigure}[t]{\spacerDET\textwidth}
     \centering
    \rotatebox{90}{~~~~~SUN $\downarrow$}
\end{subfigure}
\begin{subfigure}[t]{\spacerDE\textwidth}
     \centering
     \includegraphics[width=\textwidth]{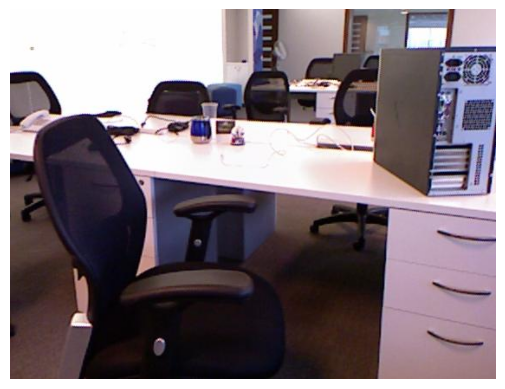}
\end{subfigure}
\begin{subfigure}[t]{\spacerDE\textwidth}
     \centering
     \includegraphics[width=\textwidth]{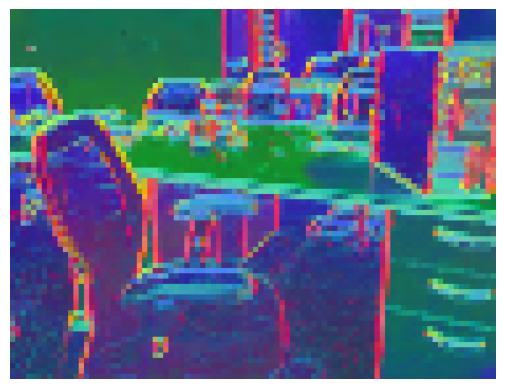}
\end{subfigure}
\begin{subfigure}[t]{\spacerDE\textwidth}
     \centering
     \includegraphics[width=\textwidth]{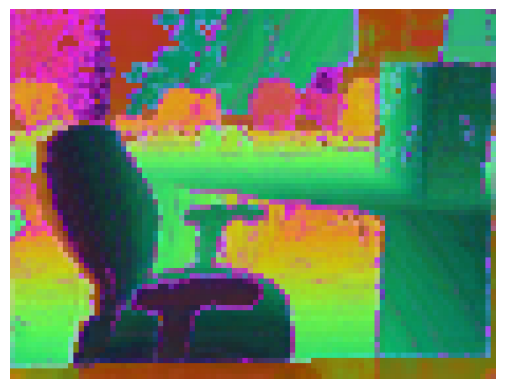}
\end{subfigure}
\begin{subfigure}[t]{\spacerDE\textwidth}
     \centering
     \includegraphics[width=\textwidth]{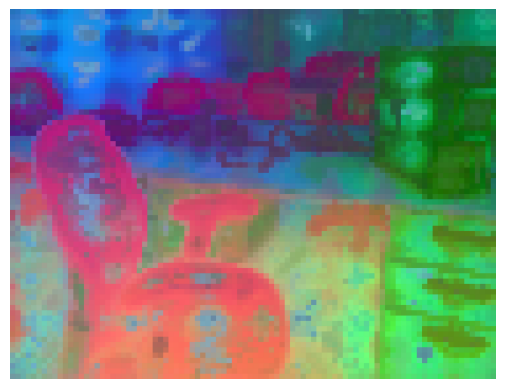}
\end{subfigure}
\begin{subfigure}[t]{\spacerDE\textwidth}
     \centering
     \includegraphics[width=\textwidth]{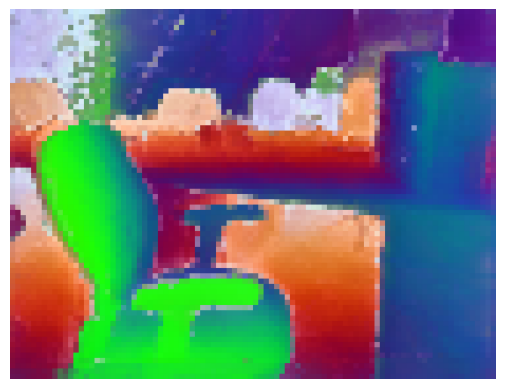}
\end{subfigure}

\hfill
\begin{subfigure}[t]{\spacerDET\textwidth}
     \centering
    \rotatebox{90}{~$\downarrow$Void}
\end{subfigure}
\begin{subfigure}[t]{\spacerDE\textwidth}
     \centering
     \includegraphics[width=\textwidth]{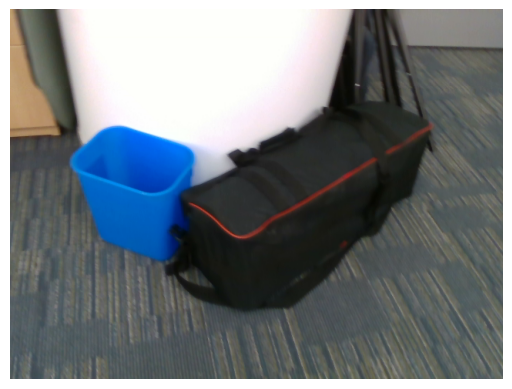}
     \caption{RGB Input \\  }
\end{subfigure}
\begin{subfigure}[t]{\spacerDE\textwidth}
     \centering
     \includegraphics[width=\textwidth]{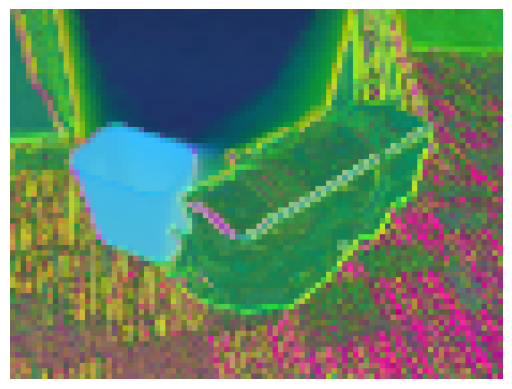}
     \caption{Ours RGB embed}
\end{subfigure}
\begin{subfigure}[t]{\spacerDE\textwidth}
     \centering
     \includegraphics[width=\textwidth]{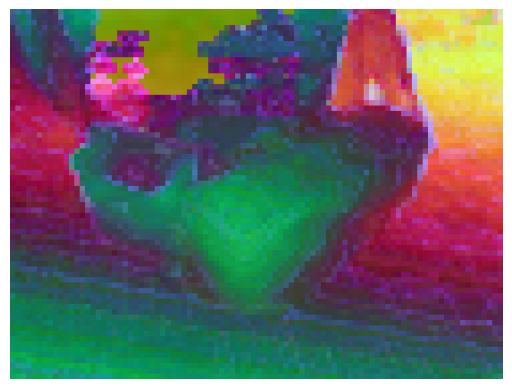}
     \caption{Ours Depth embed}
\end{subfigure}
\begin{subfigure}[t]{\spacerDE\textwidth}
     \centering
     \includegraphics[width=\textwidth]{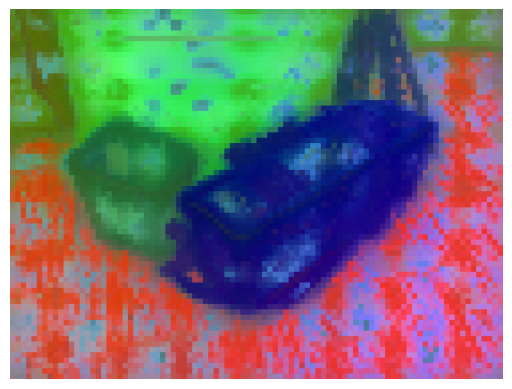}
     \caption{RGB DINOv2~\cite{DINOv2}}
\end{subfigure}
\begin{subfigure}[t]{\spacerDE\textwidth}
     \centering
     \includegraphics[width=\textwidth]{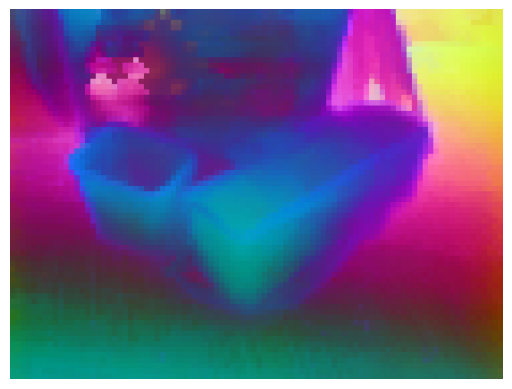}
     \caption{depth embed (ours+\cite{DINOv2})}
\end{subfigure}

\caption{\textbf{Visualization of Embeddings:} Following DINOv2 we use PCA to reduce embedded attention maps to three channels (RGB). 
}
\label{fig:vit_embs}  
\end{figure*}

\section{Results on downstream tasks}
\begin{table*}[h]
  \centering
  \caption{\textbf{Depth Completion (RMSE)}: We test the metric depth understanding of our depth adapters on related unseen depth completion datasets and rank against abs. SOTA tailored depth completion models.}
  \label{tab:depthcompletion}
  \begin{tabular}{@{}lr | c | c | c | c | c | c @{}}
    \toprule
     \multicolumn{1}{c}{}&Dataset$\rightarrow$&\multicolumn{2}{c|}{Kitti} &\multicolumn{3}{c}{Void} &NYU-V2\\
     \multicolumn{1}{c}{}&Data Split$\rightarrow$&Test&Valid&150&500&1500&500pt\\
     \multicolumn{1}{c}{}&Density$\rightarrow$&\multicolumn{2}{c|}{5\%}&0.05\%&0.15\%&0.5\% & 0.7\% \\
     \multicolumn{1}{c}{}&Max, Depth$\rightarrow$&\multicolumn{2}{c|}{80m, 16.12$\pm$11.86}&\multicolumn{3}{c}{5m, 1.65$\pm$.99}& 20m, 4.1$\pm$1.5 \\
    &Abs. SOTA$\rightarrow$&678.1~\cite{dmd3c}&-&118.5~\cite{omnidc} & 97.7~\cite{omnidc} & 74.8~\cite{omnidc} & 111.8~\cite{omnidc} \\ 
    \midrule  
    \multicolumn{2}{l|}{DMD3C (DAv2)} & \textbf{678.1} & \textbf{750.5}& 1457.8 & 1195.2 & 1021.9 & 512.3\\
    \multicolumn{2}{l|}{DMD3C (mix) } & - & 1443.6 & 666.8 & 386.8 & 210.5 & 175.9 \\ 
    \multicolumn{2}{l|}{OmniDC-v1.1} & 1154.7 & 1218.4 & \textbf{118.5} & \textbf{97.7 }& \textbf{74.8} & \textbf{111.8}\\
    \midrule
    \multicolumn{2}{l|}{DinoV2 +DA(SDP)} & 1625.06 & 1646.9 & 134.8 & 111.8 & 88.3 & 174.2 \\  
    \multicolumn{2}{l|}{~~~~ no Perlin} & - & 1501.3 & 125.6 & 102.7 & 80.2 & 155.6 \\
    \multicolumn{2}{l|}{~~~~ no PreTrained} & 1695.51 &1698.9 & 297.8 & 244.2 & 161.9 & 362.3\\ 
    \multicolumn{2}{l|}{~~~~ w. Norm} & - & 6018.2 & 1571.8 & 1567.1 & 1543.3 & 2785.2 \\    
    \multicolumn{2}{l|}{~~~~ w. DinoV3} & - & 1703.6 & 220.3 & 184.4 & 145.3 & 243.1 \\  
    \multicolumn{2}{l|}{~~~~ w. EVA02} & 1626.22 & 1649.5 & 195.5 & 160.2 & 117.2 & 366.5\\

    \bottomrule
  \end{tabular}    
\end{table*}
\label{sec:resuls}
We evaluate the transferability of our RGBD encoder across various downstream tasks. Following DINOv2, we fine-tune at a higher input resolution (448 px) to better adapt to larger images. We denote our depth-adapter variants as:
(i) \texttt{+Ours}: DA for the SDP implementation with $c{=}32$ and $\max(D)=15\,\mathrm{m}$;
(ii) \texttt{+Ours (no Perlin)}: identical to \texttt{+Ours} but trained without Perlin noise;
(iii) \texttt{+Ours (Norm)}: identical to \texttt{+Ours} with depth normalization of $5\pm5$.
We adopt GSDP with $c{=}32$ as it is computationally lighter while maintaining accuracy close to $c{=}64$. The configuration $\max(D)=15\,\mathrm{m}$ and normalization $5\pm 5$ are suitable for most downstream RGBD tasks.
For target domains with depths exceeding $\max(D)=15\,\mathrm{m}$:
(a) Depth completion on KITTI~\cite{kitti}, NYU-Depth V2~\cite{nyuDepthV2}, and Cityscapes~\cite{CityScapes}: we rescale the input and prediction at inference time to recover metric depth relative to the training range. 
(b) Semantic segmentation on the depth maps generated using UniDepth~\cite{unidepth} in ADE20K~\cite{Ade20K} and COCO-Stuff~\cite{CocoStuff}: since the absolute scale is negligible in segmentation, we rescale depths that exceed $\max(D)=15\,\mathrm{m}$ to remain within the training range. Alternatively, a new $\max(D)>15\,\mathrm{m}$ can be established that spans the entire target domain to preserve metric depth ratios, which a downstream task head can decode into metric distances. However, increasing the max depth does stretch the frequency density and increase their range (see Fig.~\ref{fig:SDP} for illustration).

\textbf{We compare our depth adapter} in terms of its ability to adapt to a large array of seen and unseen datasets across multiple disciplines based on the following points in decreasing priority: 1) the generalizability of each pretraining technique; 2) relative gains between RGB and RGBD encoders; 3) the relative gains of the generalized SOTA RGBD encoders (Multi-Modal, RGBD segmentation, and Depth Completion); 4) the relative gains of SOTA depth aware RGB encoders; 5) the absolute (abs.) SOTA (including supervised and tailored task specific methods) for a general classification of our vanilla non-finetuned encoders. 

\textbf{Our experiments} cover a wide range of related work with respect to self-supervised pretraining, generalized RGB encoder, depth completion, depth aware encoders, RGBD encoder, and multi-modal encoders. To give a complete, fair, and comprehensive comparison between all those entities, we therefore train all downstream tasks of related work with existing implementations and models using our own training pipeline and with commonly adaptable decoder heads. The quantity of experiments restricts our studies to a relatively short training schedule with limited computational resources, enabling a comprehensive comparison with repetitions to ensure the statistical significance of our results. We repeated each experiment three times to convergence and report the maximum (the standard deviation appears negligible). Unfortunately, the SOTA RGBD encoders OmniSegmentor~\cite{OmniSegmentor}, OmniVec~\cite{OmniVec}, Omnivec2~\cite{OmniVec2}, and RaSim~\cite{rasim} have not shared their implementations and models, hindering us from including them in our fair comparison with the same settings. In our following result tables, we report relative gains \textcolor{green}{green} and drops \textcolor{red}{red}. \tiny{\eye}\small seen in DA-training, $\star$ generated depth from RGB with UniDepth~\cite{unidepth}.  We mark our depth adapter with +DA.

\subsection{Depth Completion}
\label{sec:depthCompletion}
In Table~\ref{tab:depthcompletion} we summarize our depth completion results. Here we observe that Omnivore does not yield a precise pretrained metric depth understanding. These results further show that SDP yields better precision than norm-based methods on sparse input. We test on the commonly used Lidar-based Kitti~\cite{kitti} and indoor Realsense-based Void~\cite{VoidDataset} datasets. Both sets have not been seen during our pretraining. We downscale the depth input to $15$~m with a sample wise scalar if exceeded. The prediction is then rescaled with that scalar to fit the original depth distribution. Thus, our SDP can process any depth. However, stretching the frequencies does affect the precision.

\begin{table*}[h]
  \centering
  \caption{\textbf{Segmentation (mIoU)}: We train a Mask2Former~\cite{mask2former} (M2F) segmentation head on top of the frozen non-finetuned backbones (ViT-B or similar) with a relative short schedule to benchmark their of the shelf transferability.}
  \label{tab:segmentation}
  \begin{adjustbox}{width=\textwidth, center}
  \begin{tabular}{l | lr | c | c | c | c | c}
    \toprule
    \multicolumn{2}{c}{}&Dataset$\rightarrow$& NYU\cite{nyuDepthV2}\tiny{\eye} & SUN.\cite{SUNRGBD} & Citys.\cite{CityScapes}\tiny{\eye} & ADE.\cite{Ade20K}$\star $& CocoS.\cite{CocoStuff}$\star$\\
    \multicolumn{2}{c}{}&Depth$\rightarrow$& \small{4.1$\pm$1.5@86\%}&\small{2.3$\pm$0.7@87\%}&\small{27.7$\pm$26.0@78\%}&\small{13.1$\pm$7.0@100\%}&\small{10.2$\pm$6.0@100\%}\\    
    \multicolumn{2}{c}{ }&Abs. SOTA$\rightarrow$&63.6\cite{OmniVec2}&54.6\cite{GeminiFusion}&87.4\cite{SERNET-Former}&62.8\cite{BEiT3}&53.4\cite{EVA}\\
    \multicolumn{2}{c}{}&Original M2F$\rightarrow$&-&-&79.4&57.7&-\\
    \midrule
    \parbox[c]{3mm}{\multirow{6}{*}{\rotatebox[origin=c]{90}{\small{\scriptsize{Depth Aware RGB}}}}} 
    &\multicolumn{2}{l|}{FiT3D} & 46.6 & 49.5 & 68.6 & 40.5 & 34.7 \\
    &\multicolumn{2}{l|}{Dune} & 49.0 & 51.8 & 68.2 & 42.9 & 35.9\\
    &\multicolumn{2}{l|}{CroCo V2} & 24.3 & 36.4 & 55.3 & 30.2 & 23.2 \\
    &\multicolumn{2}{l|}{MASt3R (ViT-L)} & 38.5 &  44.2& 65.3 & 33.1 & 24.7\\
    &\multicolumn{2}{l|}{DepthAnythinV2} & 52.0 & 54.8 & 74.5 & 48.7 & 43.9\\
    &\multicolumn{2}{l|}{~~~~ +DinoV2} & 53.5(\textcolor{green}{1.5}) & 55.3(\textcolor{green}{0.5}) & 75.9(\textcolor{green}{1.4}) & 49.7(\textcolor{red}{2.0}) & 45.7(\textcolor{green}{1.8})\\       
    \midrule         
    \parbox[c]{3mm}{\multirow{4}{*}{\rotatebox[origin=c]{90}{RGBD}}}&\multicolumn{2}{l|}{DFormerV1-pre} & 42.2 & 43.9 & 74.6 & 44.7 & 40.8\\
    &\multicolumn{2}{l|}{DFormerV2-pre} & 33.5 & 40.3 & 71.0 & 41.4 & 38.7\\
    &\multicolumn{2}{l|}{OmniDC-pre} & 31.6 & 38.4 & 58.3 & 36.9 & 11.2 \\
    &\multicolumn{2}{l|}{DMD3C-pre} & 31.0 & 38.3 & 72.8 & 36.3 \textbf{}& 29.9\\
    \midrule  
    \parbox[c]{3mm}{\multirow{4}{*}{\rotatebox[origin=c]{90}{\scriptsize{Multi-Modal}}}}&\multicolumn{2}{l|}{Omnivore} & 43.5 &46.5 & 71.8 & 45.9 & 42.7 \\
    &\multicolumn{2}{l|}{~~~~ +Depth} & 42.3(\textcolor{red}{1.2}) & 40.9(\textcolor{red}{5.6}) & 61.0(\textcolor{red}{10.8}) & 44.3(\textcolor{red}{1.6}) & 41.9(\textcolor{red}{0.8}) \\
    &\multicolumn{2}{l|}{MultiMAE} & 29.7 & 36.8 & 58.6 & 30.8 & 26.0\\
    &\multicolumn{2}{l|}{~~~~ +Depth} & 36.7(\textcolor{green}{7.0}) & 41.0(\textcolor{green}{4.2}) & 67.2(\textcolor{green}{8.6}) & 41.1(\textcolor{green}{10.3}) & 36.7(\textcolor{green}{10.7}) \\
    \midrule
    \parbox[c]{3mm}{\multirow{11}{*}{\rotatebox[origin=c]{90}{\scriptsize{SOTA Encoders +Ours}}}}&\multicolumn{2}{l|}{DA(No pre-training)}  & 22.4 & 32.1 & 57.2 & 29.5 & 23.8 \\
    
    &\multicolumn{2}{l|}{~~~~ +DA(SDP)}  & 29.6(\textcolor{green}{7.2}) & 38.0(\textcolor{green}{5.9}) & 60.6(\textcolor{green}{3.4}) & 36.6(\textcolor{green}{7.1}) & 29.7(\textcolor{green}{5.9}) \\

    &\multicolumn{2}{l|}{EVA-02} & 30.1 & 39.1 & 48.4 & 32.7 & 28.7 \\
      &\multicolumn{2}{l|}{~~~~ +DA(SDP)} & 38.1(\textcolor{green}{8.0}) &44.1(\textcolor{green}{5.0}) & 66.5(\textcolor{green}{18.1}) & 43.5(\textcolor{green}{10.8}) & 39.6(\textcolor{green}{10.9}) \\

      &\multicolumn{2}{l|}{DINOv2} & 52.7 &53.8 & 74.4 & 51.7 & 44.6 \\ 
      &\multicolumn{2}{l|}{~~~~ +DA(SDP)} & 54.8(\textcolor{green}{2.1}) &\textbf{56.1}(\textcolor{green}{2.3}) & \textbf{77.7}(\textcolor{green}{3.3}) & 53.4(\textcolor{green}{1.7}) &\textbf{49.9}(\textcolor{green}{5.3}) \\ 
      &\multicolumn{2}{l|}{~~~~ +DA(Norm)} & \textbf{55.1}(\textcolor{green}{2.4}) & 55.8(\textcolor{green}{2.0}) & 77.1(\textcolor{green}{2.7}) & \textbf{53.6}(\textcolor{green}{1.9}) & 49.5(\textcolor{green}{4.9})\\ 
      &\multicolumn{2}{l|}{~~~~ +DA(No Perlin)} & 54.6 & 55.3 & 77.5 & 54.0 & 49.9\\ 

      &\multicolumn{2}{l|}{DINOv3} & 53.0 & 54.6 & 72.6 & 48.9 & 43.0 \\ 
      &\multicolumn{2}{l|}{~~~~ +DA(SDP)} & 54.5(\textcolor{green}{1.5}) & 55.2(\textcolor{green}{0.6}) & 76.5(\textcolor{green}{3.9}) & 54.7(\textcolor{green}{5.8})  & 49.7(\textcolor{green}{6.7}) \\ 
    \bottomrule
  \end{tabular}  
  \end{adjustbox}
\end{table*}

\subsection{Segmentation}
\label{sec:segmentation}
Using MMSegmentation~\cite{mmseg}, we train a commonly used Mask2Former~\cite{mask2former} segmentation head on top of our frozen encoders. In Table~\ref{tab:segmentation}, we summarize our segmentation results in the original resolution, where we are able to achieve absolute SOTA results of 56.05 mIoU on SUN-RGBD. Please note that the depth data for CocoS.~\cite{CocoStuff} and Ade.~\cite{Ade20K} is generated using monocular depth estimation, showing our methods effectiveness on RGB-only datasets. Unlike our results, we observe again that the raw pretrained Omnivore gets negatively affected when using depth. We select Mask2Former as it is a general method that works well for any encoder or dataset. We use a resolution of 518 for our ViT/14 (patch-size) and 512 for Omnivore/16 with a batch size of 16, and train for 20k steps on NYUv2~\cite{nyuDepthV2} and SUN-RGBD~\cite{SUNRGBD}, and 30k steps on the larger ADE20k~\cite{Ade20K} and COCO-Stuff~\cite{CocoStuff} datasets. We again use ViT fusion layers [3, 6, 9, 12] for the gradual decoding following~\cite{DPT, DINOv2}. We use interpolation to reshape the ViT feature maps into the hierarchical shape required by Mask2Former. For data augmentation, we employ random crops, color jitters, flips, grayscale, solarization, and Gaussian blurring.

\begin{table*}[h]
  \centering
    \caption{\textbf{Pose estimation}: We train the fast multi-object PoET(GT)~\cite{PoET} 6D ojbect pose estimation head on top of the frozen encoders without any further 3D postprocessing or encoder finetuning. We compare our DA adapter vs. their RGB baselines, the SOTA RGBD encoder Omnivore~\cite{OmniVec}, and rank them w.r.t the abs. SOTA. We report the commonly used metric per dataset.}
  \label{tab:poseestimation}
  \begin{adjustbox}{width=\textwidth, center}
  \begin{tabular}{@{}l| lr | c | c | c | c | c | c | c @{}}
    \toprule
     \multicolumn{2}{c}{}&Dataset$\rightarrow$&\multicolumn{2}{c|}{YCBV\cite{PoseCNN}\tiny{\eye}}& \multicolumn{2}{c|}{HB\cite{HomeBrew}} & \multicolumn{3}{c}{LM-O\cite{LM-O}\tiny{\eye}}\\
     \multicolumn{2}{c}{}&Depth$\rightarrow$& \multicolumn{2}{c|}{\small{1.2$\pm$0.5@82\%}}&\multicolumn{2}{c|}{\small{1.1$\pm$0.4@90\%}}&\multicolumn{3}{c}{\small{1.3$\pm$0.9@84\%}}\\
    \multicolumn{2}{c}{}& Metric$\rightarrow$&AUC\cite{PoseCNN}&AR\cite{BoP23}&AUC\cite{PoseCNN}&AR\cite{BoP23}$\dagger$&AUC\cite{PoseCNN}&ADD(-S)\cite{LineMod}&AR\cite{BoP23}\\
    \multicolumn{2}{c}{}&Abs. SOTA$^\ddagger$$\rightarrow$& 95.4~\cite{Enhanced6D} & 93.8~\cite{BoP23}& - &  95.6~\cite{BoP23} & - &  89.6~\cite{hipose} &80.5~\cite{BoP23}\\
    \multicolumn{1}{c}{}&&PoET~\cite{PoET}$\rightarrow$& 92.8 & 68.3 & - & - & - & 36.8 & - \\
    \midrule
    \parbox[c]{3mm}{\multirow{6}{*}{\rotatebox[origin=c]{90}{\scriptsize{Depth Aware RGB}}}}
     & \multicolumn{2}{l|}{FiT3D} & 70.3 & 55.1 & 46.5 & 32.3 & 52.9 & 19.5 & 39.0 \\
     & \multicolumn{2}{l|}{Dune} & 72.2 & 57.5 & 50.8 & 37.8 & 60.4 & 23.0 & 41.8 \\
     & \multicolumn{2}{l|}{CroCoV2} & 61.9& 43.2 & 35.8 & 20.2 & 53.9 & 18.0 & 34.2  \\
     & \multicolumn{2}{l|}{MASt3R (ViT-L)} & 66.2 & 49.6 & 39.6 & 25.9 & 63.0 & 26.4 & \textbf{43.4} \\
     & \multicolumn{2}{l|}{DepthAnythingV2} &  71.7& 57.0  & 48.7 & 34.6 & 54.9 & 21.1 & 41.7 \\
     & \multicolumn{2}{l|}{~~~~~ +DinoV2 } & 72.5 & 59.3 & 40.3 & 33.3 & 48.7 & 17.3 & 38.8  \\
     \midrule
    \parbox[c]{3mm}{\multirow{4}{*}{\rotatebox[origin=c]{90}{RGBD}}}&\multicolumn{2}{l|}{DFormerV1-pre}& 63.3& 37.7 & 41.1 & 25.0 & 39.1 & 13.0 & 27.6  \\
     &\multicolumn{2}{l|}{DFormerV2-pre}& 59.4& 38.1 & 15.8 & 14.9 & 41.6 & 11.1 & 33.8 \\
     &\multicolumn{2}{l|}{OmniDC-pre}& 39.9 & 15.8 &20.2 & 5.8 & 34.0 & 7.2 & 16.3  \\
      &\multicolumn{2}{l|}{DMD3C-pre}& 46.2 & 24.2 & 20.6 & 8.9 & 40.4 &  11.1 & 20.6 \\   
    \midrule
    \parbox[c]{3mm}{\multirow{4}{*}{\rotatebox[origin=c]{90}{\scriptsize{Multi-Modal}}}}&\multicolumn{2}{l|}{Omnivore}& 62.6& 45.9 & 46.5 & 23.3 & 53.9 & 18.2 & 37.4 \\
    &\multicolumn{2}{l|}{~~~~+ Depth} & 55.0(\textcolor{red}{7.6}) & 37.8(\textcolor{red}{8.1}) & 41.5(\textcolor{red}{5.0}) &14.6(\textcolor{red}{8.7}) & 45.7(\textcolor{red}{8.2}) & 16.4(\textcolor{red}{1.8}) & 31.1(\textcolor{red}{6.3}) \\
    &\multicolumn{2}{l|}{MultiMAE}& 64.3 & 45.3 & 37.4 & 20.9 & 53.6 & 17.2 & 35.1  \\
    &\multicolumn{2}{l|}{~~~~ + Depth} & 69.9(\textcolor{green}{5.6})& 45.7(\textcolor{green}{0.4}) & 52.1(\textcolor{green}{14.7}) & 35.6(\textcolor{green}{14.7}) & 47.5(\textcolor{red}{6.1}) & 12.0(\textcolor{red}{5.2}) & 26.5(\textcolor{red}{8.6}) \\ 
    \midrule
    \parbox[c]{3mm}{\multirow{8}{*}{\rotatebox[origin=c]{90}{\scriptsize{SOTA Encoders +Ours}}}}&\multicolumn{2}{l|}{EVA-02}& 64.3 & 47.3 & 38.2 & 22.2 & 62.2 & 24.6 & 41.1 \\
    &\multicolumn{2}{l|}{~~~~ +DA(SDP)} &  76.4(\textcolor{green}{20.3})&  54.9(\textcolor{green}{7.6s})& 61.3(\textcolor{green}{23.1}) & 51.4(\textcolor{green}{28.8}) & 62.2(\textcolor{gray}{0.0})& 26.3(\textcolor{green}{1.7})& 36.3(\textcolor{red}{4.8})\\ 
    &\multicolumn{2}{l|}{DINOv2}& 70.6& 58.3 & 50.1 & 38.8 & 51.9 & 17.8 & 38.7 \\
    &\multicolumn{2}{l|}{~~~~ +DA(SDP)} & \textbf{84.1}(\textcolor{green}{13.5})& \textbf{68.7(\textcolor{green}{10.4})} &\textbf{68.9}(\textcolor{green}{18.8})&\textbf{54.2}(\textcolor{green}{15.4}) & 64.5(\textcolor{green}{12.6}) &29.8(\textcolor{green}{12.0})& 40.9(\textcolor{green}{2.2}) \\  
    &\multicolumn{2}{l|}{~~~~ +DA(Norm)} & 83.4(\textcolor{green}{12.8}) & 67.6(\textcolor{green}{9.3}) &63.8(\textcolor{green}{13.7}) &46.1(\textcolor{green}{7.3}) & 64.5(\textcolor{green}{12.6}) &27.3(\textcolor{green}{9.5})& 40.2(\textcolor{green}{1.5}) \\
    &\multicolumn{2}{l|}{~~~~ +DA(No Perlin)} & 83.6(\textcolor{green}{13.0})& 67.7(\textcolor{green}{9.4}) &  67.7(\textcolor{green}{17.6}) & 51.2(\textcolor{green}{12.4}) & 64.4(\textcolor{green}{12.4}) & 29.7(\textcolor{green}{11.9}) & 41.3(\textcolor{green}{2.6}) \\
    &\multicolumn{2}{l|}{DINOv3}& 69.3 & 53.2 & 44.5 & 32.4 & 47.2 & 15.5 & 38.1 \\
    &\multicolumn{2}{l|}{~~~~ +DA(SDP)}& 82.4(\textcolor{green}{13.1}) & 62.2(\textcolor{green}{9.0}) & 66.1(\textcolor{green}{21.6}) & 53.7(\textcolor{green}{21.3}) & \textbf{65.6}(\textcolor{green}{18.4}) & \textbf{31.2}(\textcolor{green}{15.7}) & \textbf{42.3}(\textcolor{green}{4.2}) \\
    \bottomrule
  \end{tabular}  
  \end{adjustbox}
\end{table*}

\begin{figure}[h]
    \centering
    \includegraphics[width=0.6\linewidth]{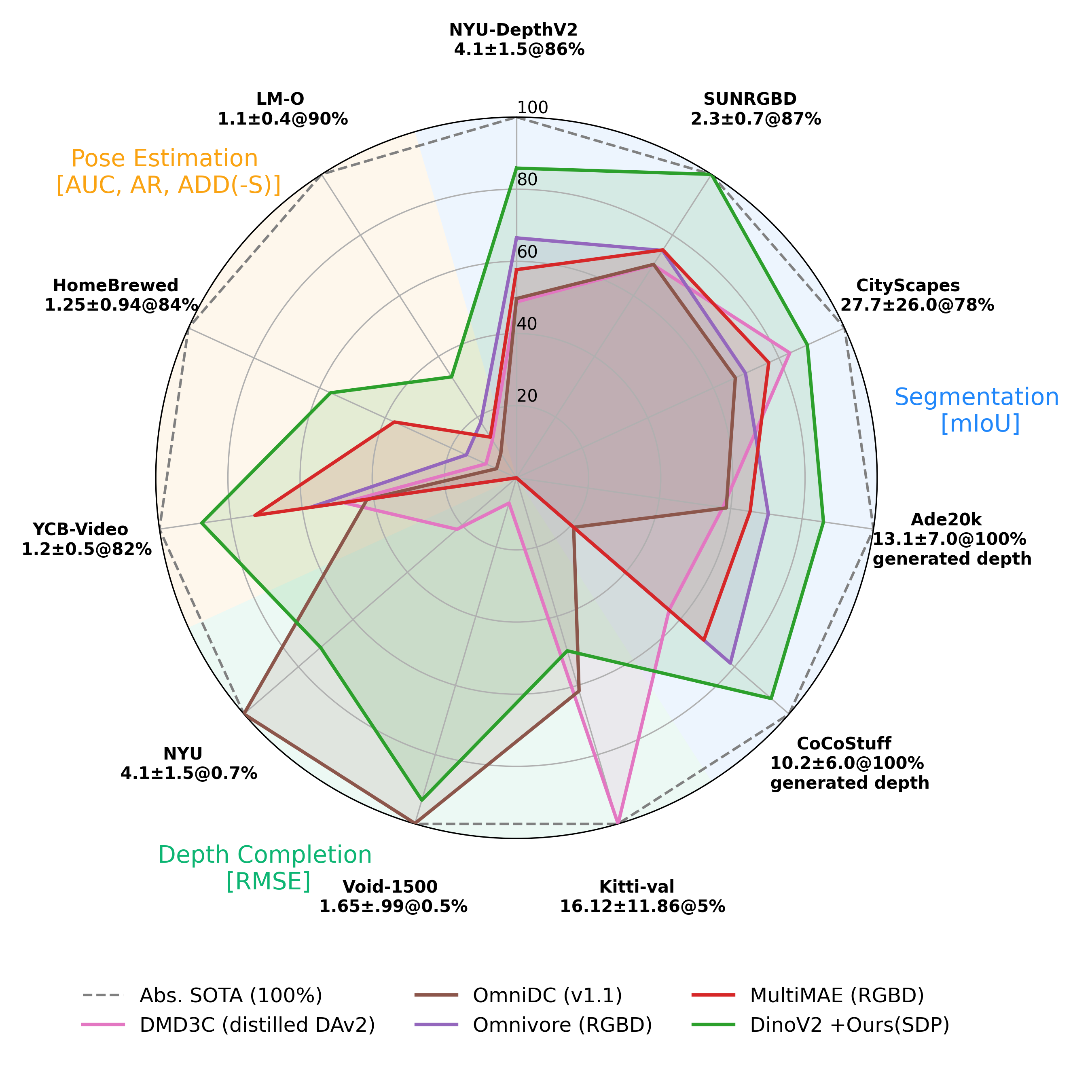}
    \caption{\textbf{Relative in \% to the abs. SOTA:} Ours +DinoV2~\cite{DINOv2} yields better generalized results w.r.t SOTA models (MultiMAE~\cite{multimae}, Omnivore~\cite{OmniVore}) on RGBD downstream tasks in Pose estimation and Segmentation, while compatible with specialized depth completion models (DMD3C~\cite{dmd3c}, Omni-DC~\cite{omnidc}).}
    \label{fig:radar_sota}
\end{figure}

\def\spacerDE{0.46}
\begin{figure*}[h!]
\centering
\begin{subfigure}[b]{\spacerDE\textwidth}
     \centering
     \includegraphics[width=\textwidth]{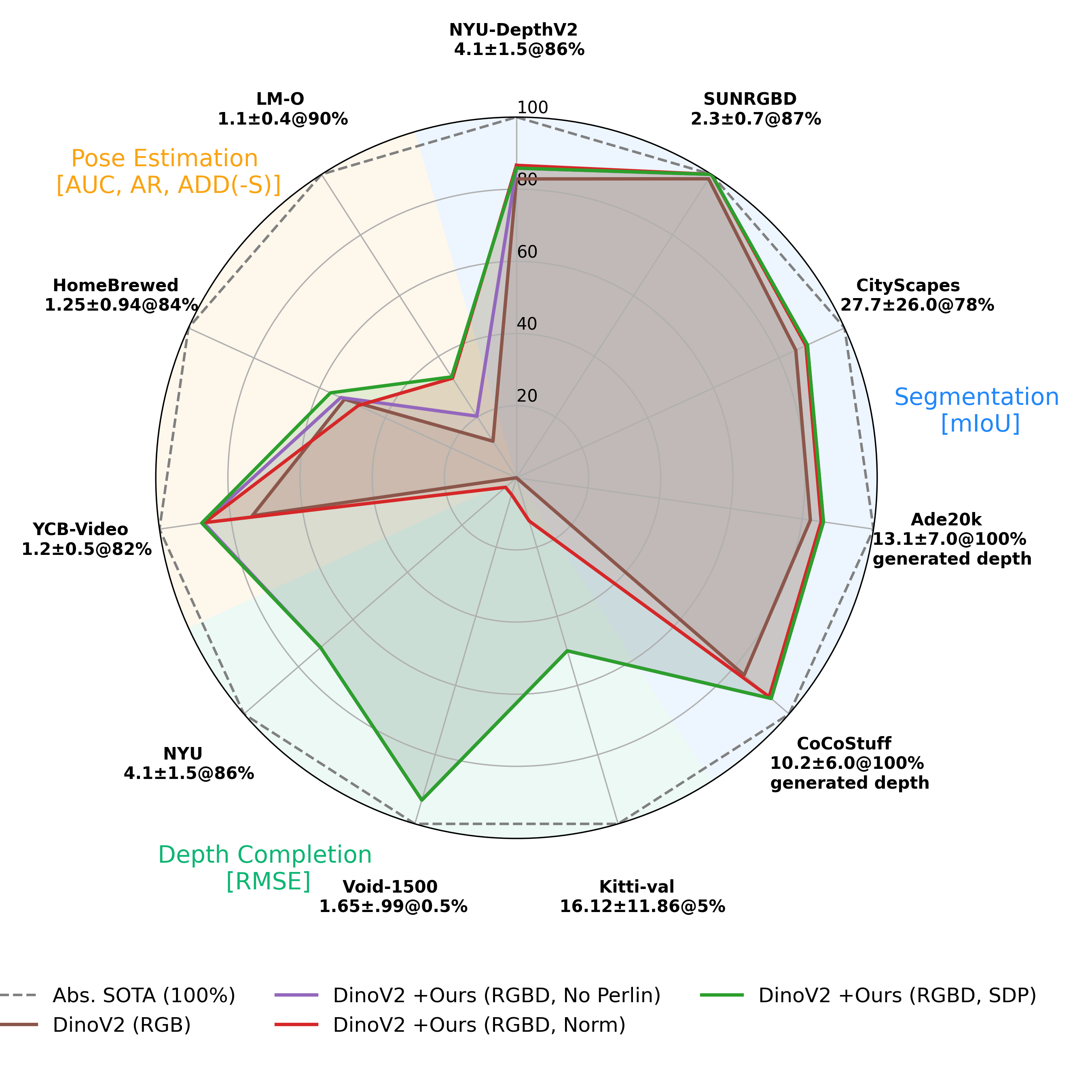}
     \caption{SDP vs. depth normalization vs. train with perlin noise.}
\end{subfigure}
\begin{subfigure}[b]{\spacerDE\textwidth}
     \centering
     \includegraphics[width=\textwidth]{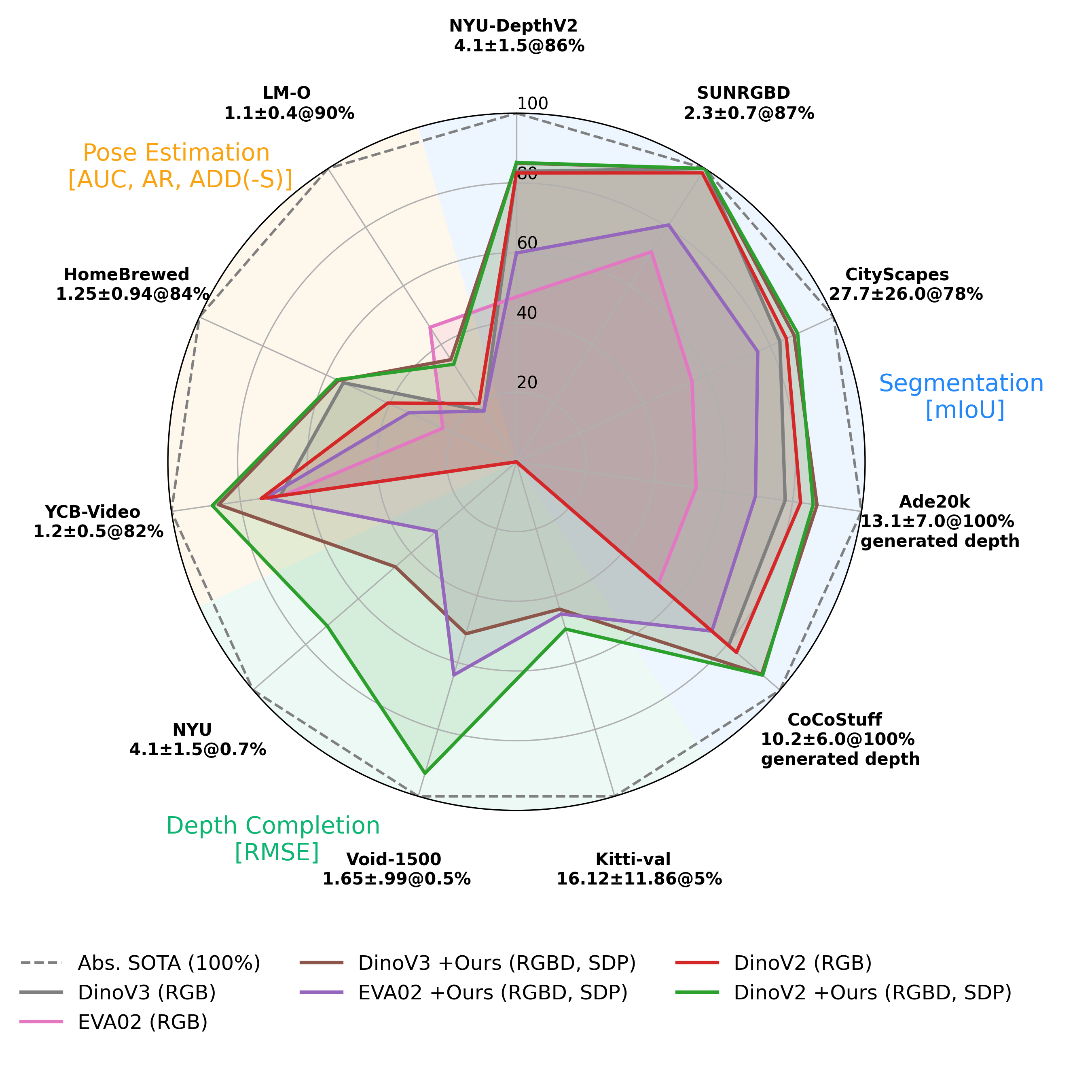}
     \caption{Our depth adapter improves multiple generalized RGB encoders.}
\end{subfigure}
\begin{subfigure}[t]{\spacerDE\textwidth}
     \centering
     \includegraphics[width=\textwidth]{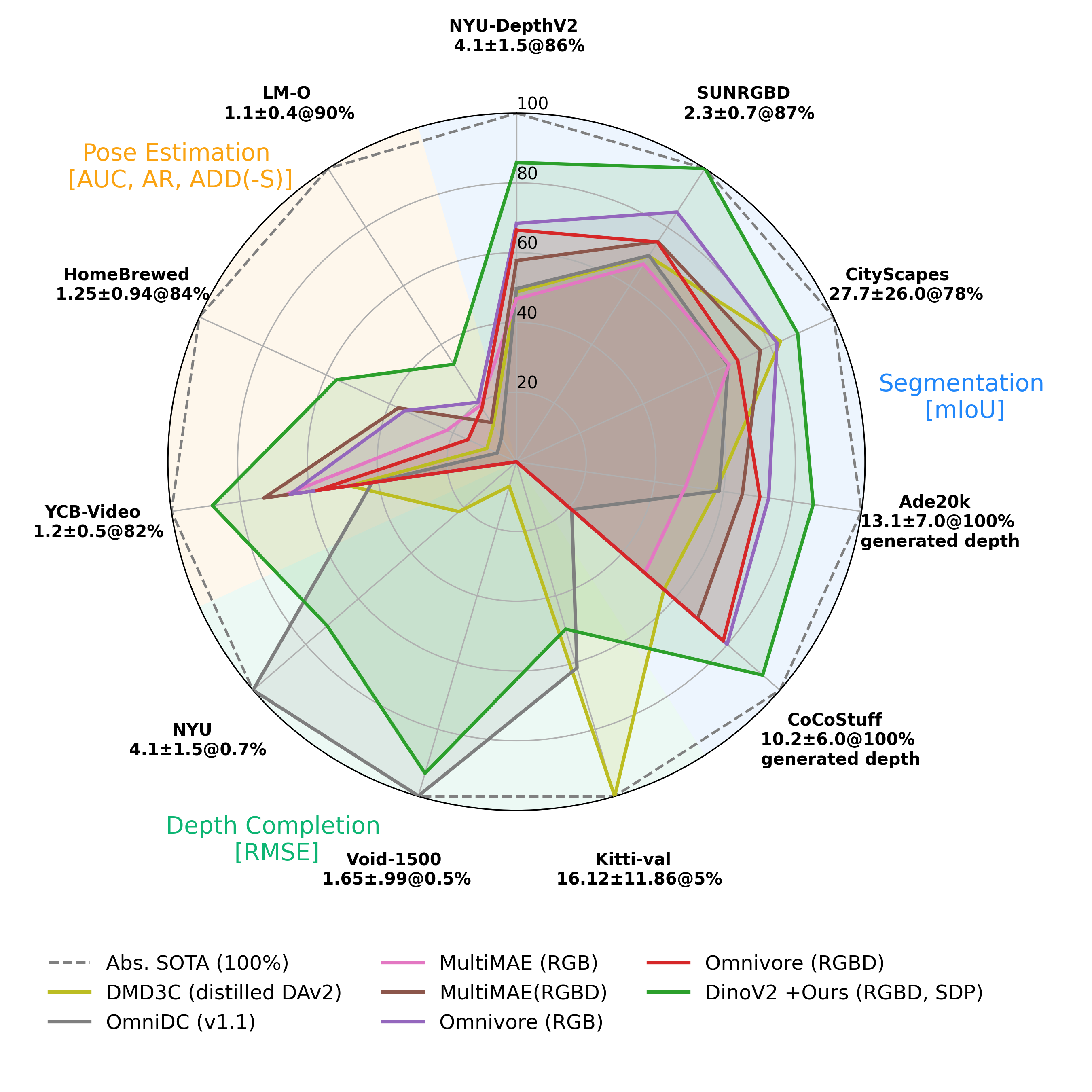}
     \caption{Ours vs. Multi-Modal, RGBD, and depth completion encoders. }
\end{subfigure}
\begin{subfigure}[t]{\spacerDE\textwidth}
     \centering
     \includegraphics[width=\textwidth]{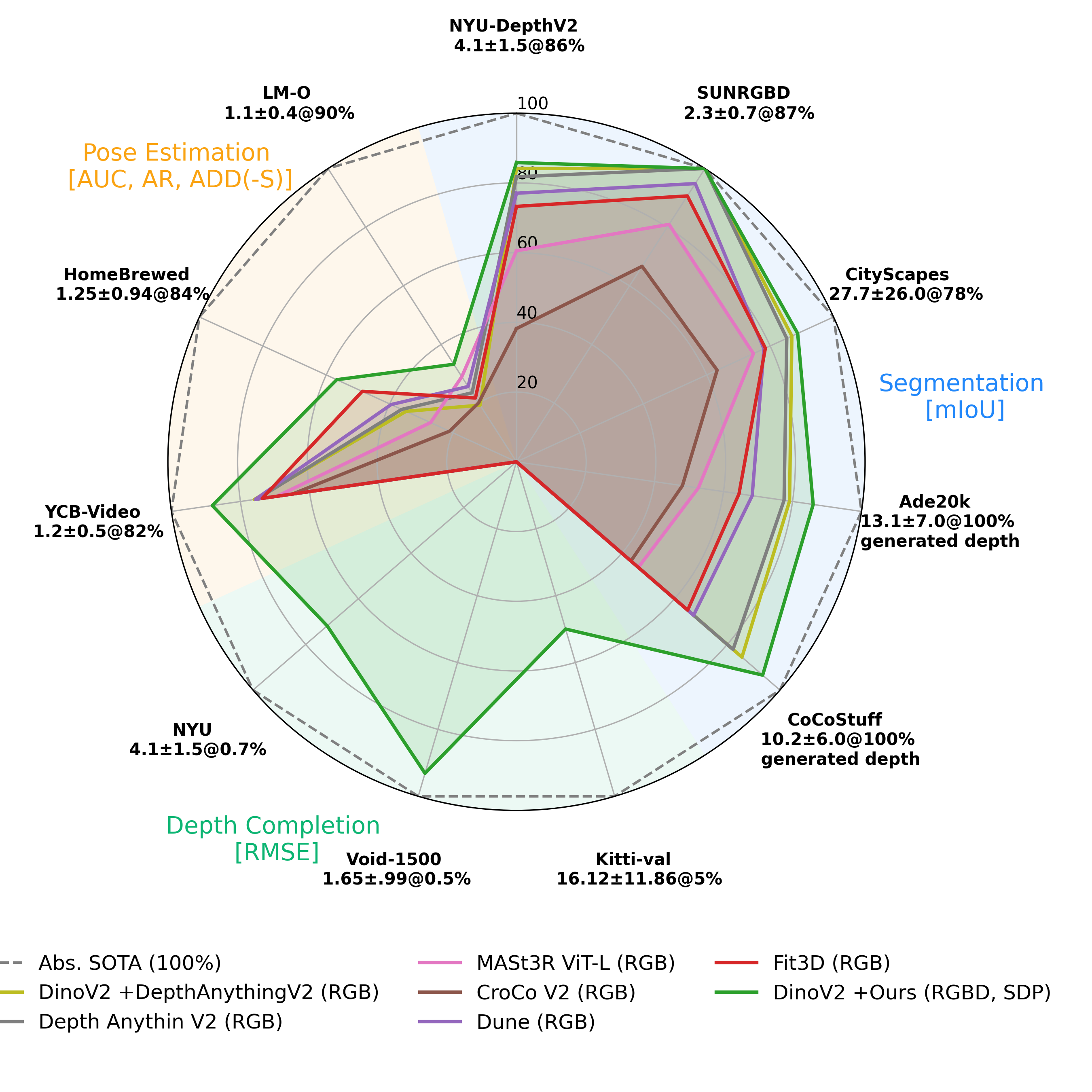}
     \caption{Ours vs. depth aware RGB encoders.}
\end{subfigure}
\caption{\textbf{Visualization of Generalization:} Raw generalizeability of frozen pretrained encoders (no finetuning) with common general adoptable lightweight decoders. We compare our results against various related generalized, task oriented, and abs. SOTA methods across a vide range of downstream tasks.}
\label{fig:radar_plots}  
\end{figure*}

\subsection{6D object pose estimation}
\label{sec:6Dposeestimation}
Note that Omnivore and MultiMAE do not address 6D pose estimation or any other disciplines besides classification and segmentation. We believe that robotic applications are most relevant for RGBD encoders. Therefore, we train a simple PoET~\cite{PoET} 6D object pose estimation head on top of our frozen encoders to highlight relative gains. PoET follows the generalized decoder architecture of Deformable-DeTR~\cite{DeformableDETR} and is applicable to any encoder. PoET works without any data-oriented tailored design or post-processing, which makes it a perfect fit to analyze and compare the encoders' generalization and adaptability. However, the absolute results of PoET are not comparable to the absolute SOTA, which are often much slower, larger, employ 3D post-processing (e.g. RaSim~\cite{rasim}), and task-specific encoder-decoder designs (e.g. RaSim~\cite{rasim}). The absolute SOTA in pose estimation mostly features single object, slow, tailored methods with 3D post processing. Ours works with $>10$ FPS independent of the dataset, while the absolute SOTA "SCFlow"~\cite{BoP23} needs $0.55$~s, $1.82$~s, and $1.70$~s per image, for YCBV, HB, and LM-O, respectively. In Table~\ref{tab:poseestimation} we summarize our results. Here, we can see the most significant impact of our DA encoders with respect to their baselines. Omnivore is again negatively affected by RGBD for all datasets but LM-O. SDP prove to be more resilient than norm-based depth encoding on average. We train a PoET~\cite{PoET} 6D object pose estimation head on top of our frozen encoder weights for 90k steps and only color augmentations with a batch size of 48 for non interpolated ViTs feature maps and batch size 16 for hierarchical feature maps (e.g. Omnivore). Both batch sizes use the full 48G VRAM of a Nvidia RTX 6000 ADA, thus the same computational resources. We use PoET as it follows the general decoding method of Deformable DETR~\cite{DeformableDETR}, can be added on top of any encoder, and requires no post-processing or (often iterative) second stage. Following PoET we use a lr of 2e-4 for the decoder head. We train our pose estimation models for YCBV~\cite{PoseCNN} (including the 80k synthetic training samples), Homebrew~\cite{HomeBrew}, and LM-O~\cite{LM-O}, which are from the commonly used BOP (pose estimation) challenge~\cite{BoP23}. YCBV features 21 objects with less than 10 objects per scene. The Sim2real challenge of Homebrew has 33 objects, sometimes more than 20 objects per scene. Therefore, we extend PoET's default number of 10 decoding queries to 30. LM-O has 8 objects with the challenge of relative high occlusion.  For our pose estimation results see. Table~\ref{tab:poseestimation}. $^\dagger$For HB~\cite{HomeBrew} we use the validation Bop~\cite{BoP23} set, since the test set has no public bounding boxes, which are required by PoET(GT). 

\begin{figure}[h!]
    \centering
    \includegraphics[width=0.95\linewidth]{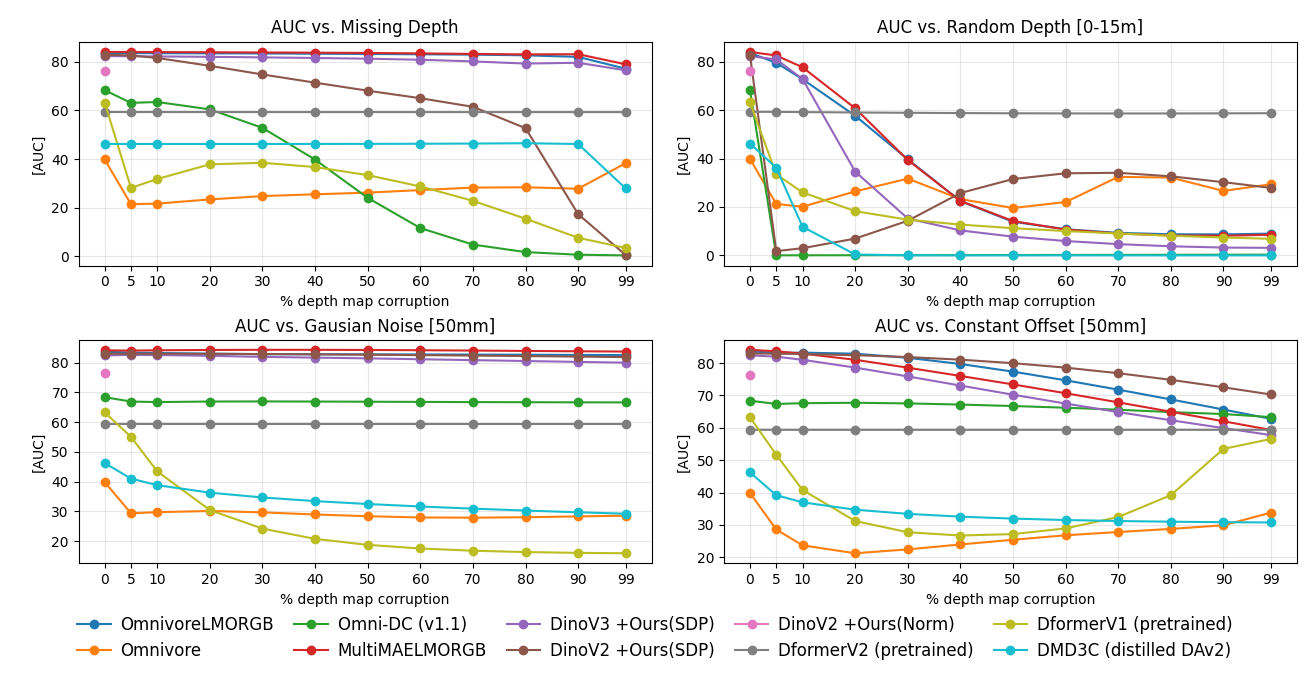}
    \caption{
    \textbf{Depth perturbations} affect the pose estimation on YCBV~\cite{PoseCNN}. For \textit{missing depth} and \textit{gaussian noise}, the models should keep a high AUC to show robustness. For \textit{constant offset} and \textit{random depth} the AUC should drop with depth misinformation to show that the models are dependent the depth input.
    }
    \label{fig:depth_noise_curves}
\end{figure}

\begin{figure}[h]
    \centering
    \includegraphics[width=0.95\linewidth]{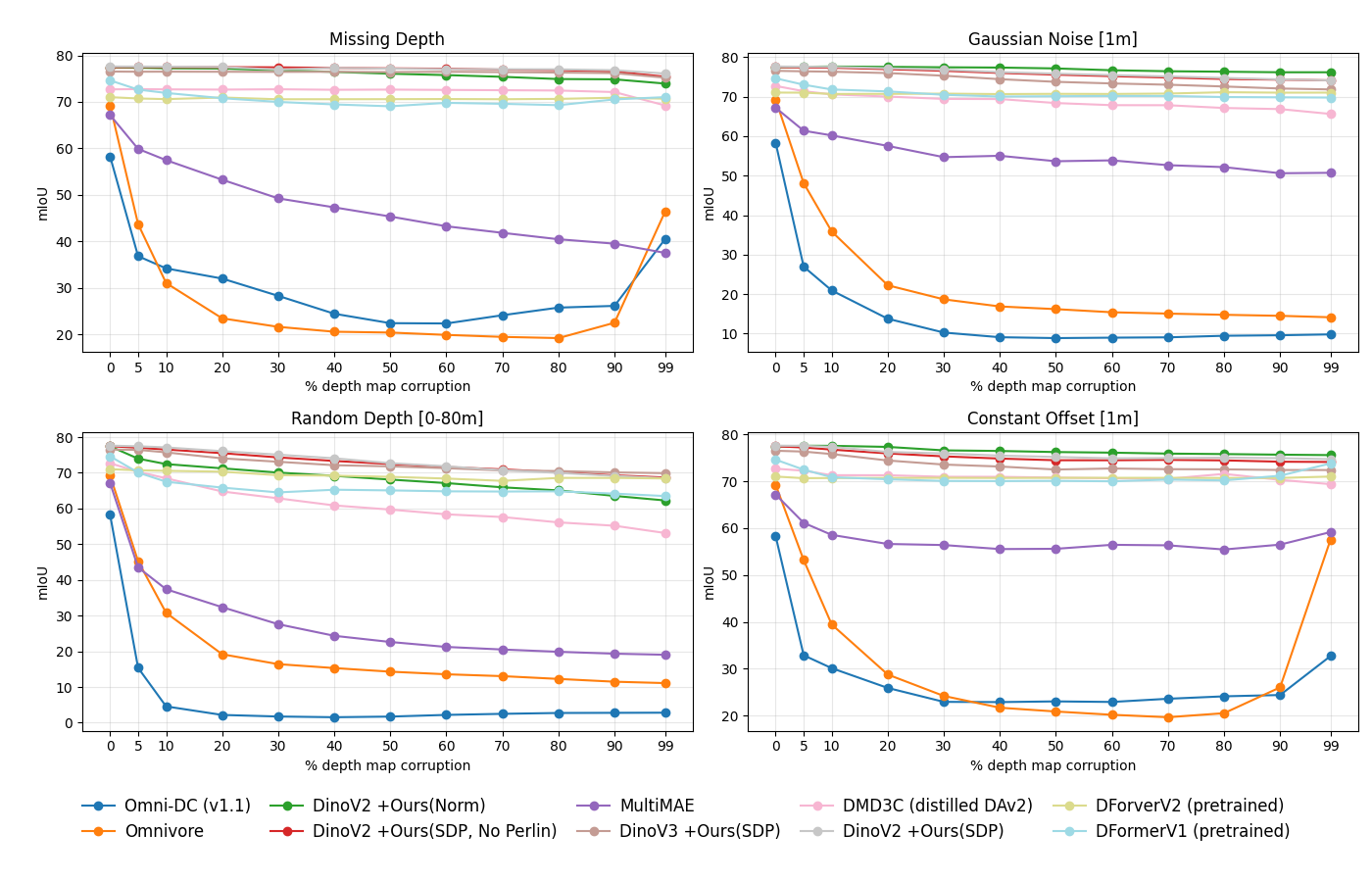}
    \caption{
    \textbf{Depth perturbations} affect the segmentation on Cityscapes~\cite{CityScapes}. For \textit{missing depth} and \textit{gaussian noise}, the models should keep a high AUC to show robustness. When the \textit{constant offset} is consistent the models approach the original results as precise metric depth is not relevant for most segmentation tasks.
    }    
    \label{fig:citysegcorruption}
\end{figure}

\section{Discussion of Generalizability} 
In Fig.~\ref{fig:radar_sota}~\&~\ref{fig:radar_plots}, we visualize the relative generalizability of our methods and the multi-modal SOTA and rank them against the absolute tailored SOTA. Please note that a direct comparison against the absolute SOTA is unfair, as these works use different encoder sizes, are task-specific tailored with fitted decoders, use more compute, finetune the encoders, and often utilize special postprocessing. In our results of Fig.~\ref{fig:depth_noise_curves}~\&~\ref{fig:citysegcorruption}, we show that SDP is more stable than norm-based depth encoding, especially in depth completion and pose estimation, where one can expect a higher importance of precise metric depth understanding than in segmentation tasks. Both the SDP and norm-based depth adapters can handle strong gaussian noise. However, norm-based encoding struggles with sparsity and random noise depth significantly more than SDP. When applying a constant offset, both methods drop accordingly but seem to still perform more stably than others. This indicates that the remaining relative depth understanding still benefits pose estimation (70.6 AUC in the RGB DinoV2 baseline). However, when misinformation in the depth noise increases, we expect the models to drop below the RGB baseline, demonstrating that they rely on consistent depth information. However, for depth normalization, we can see that the model is not doing that. This indicates that the normalized depth is mostly ignored and causes catastrophic turbulence within the encoders when exposed to noise or input density shifts. In contrast, Omnivore~\cite{OmniVore} and MultiMAE~\cite{multimae} can withstand gaussian noise; they drop significantly with missing or random depth inputs. For constant offset, we can see that the models are not significantly affected. This shows us that both models are not significantly affected by absolute metric depth. Instead, they are aware of relative depth due to their classification and MAE-based pretraining (see Sec.~\ref{sec:related_work}). SDP is shown to work compatibly better than depth normalization with sparse and noisy inputs, as shown in depth completion in Fig~\ref{fig:radar_plots} and Fig.~\ref{fig:depth_noise_curves}. Moreover, in our ablation (see Table~\ref{tab:DA_results}), we show that SDP yields a significantly more precise encoding-decoding than norm-based approaches. 

\section{Conclusion}
\label{sec:conclusion}
We present SDP and vanishing depth: a self-supervised pipeline to train generalized depth adapters for existing generalized RGB encoders. With SDP, our adapters demonstrate a general capability to address related RGBD downstream tasks without further finetuning and on various depth distributions and densities. They exceed the RGB baselines of their predecessors and generalized RGBD SOTA competitors on various downstream benchmarks. Notably, we achieve absolute SOTA with 56.05 mIoU on SUN-RGBD segmentation. In pose estimation, our adapters outperform their multi-modal or RGBD predecessors and depth-aware RGB competitors. SDP achieves a higher precision than using SOTA norm-based depth preprocessing. This is particularly evident in our depth completion and perturbation experiments. Although our norm-based encoders yield similar results to their SDP-based counterparts, we show that SDP is generally more stable and less affected by shifting depth input. This is especially important for depth inputs, which often fluctuate in test time environments. Our work aims to contribute to research for generalized image encoders, which enable generalization across various domains, distributions, densities, problem sizes, and downstream tasks. This is achieved without the need for finetuning, allowing for fast adaptation on a shared encoder for efficient and modular multi-agent systems. Here, we especially see the potential of our generic RGBD embeddings in the field of VLA and robot policy learning, which generally benefits from pretrained encoders but requires finetuning at the cost of overfitting to their training environment. We believe that SDP can be further adapted to benefit other modalities that are facing similar issues related to encoding continuous unbounded numbers.

\section*{Declaration of AI}
GenAI in the form of LLMs is used only for text formatting, grammar correction, and formulation suggestions. 

\small
\bibliographystyle{bibtex/splncs03}
\bibliography{refs}

\end{document}